\documentclass{article}

\usepackage[
    a4paper,
    left=1in,
    right=1in,
    top=1.35in,
    bottom=1.35in,
    headheight=26pt,
    headsep=28pt,
    footskip=30pt
]{geometry}
\usepackage[utf8]{inputenc}
\usepackage[T1]{fontenc}
 \usepackage{multirow} 
 \usepackage{longtable}
\usepackage[
  protrusion=true, expansion=true, final,
  factor=1100, stretch=20, shrink=20
]{microtype}

\usepackage{amsmath,amssymb,amsfonts}
\usepackage{makecell}
\usepackage{graphicx}
\usepackage{subcaption}
\usepackage{booktabs}
\usepackage{array}
\usepackage{multicol}

\usepackage[table]{xcolor}
\definecolor{groupbg}{RGB}{255,229,204}
\definecolor{gold}{RGB}{212,175,55}
\definecolor{silver}{RGB}{150,150,150}
\definecolor{bronze}{RGB}{205,127,50}
\definecolor{priorlight}{RGB}{247,248,250}

\usepackage{hyperref}
\usepackage{url}
\usepackage{cleveref}
\usepackage{natbib}

\usepackage{enumitem}
\usepackage{xspace}

\usepackage{caption}
\usepackage{setspace}
\usepackage{fancyhdr}
\fancypagestyle{withlogo}{%
  \fancyhf{}
  \fancyhead[L]{\includegraphics[height=18pt]{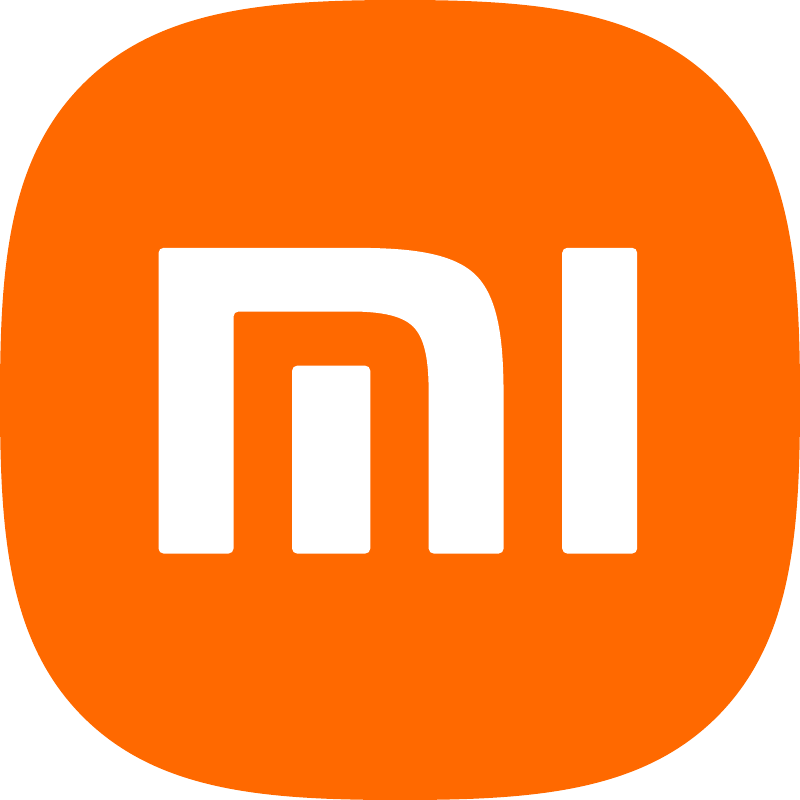}}
  \fancyhead[R]{\textsf{Xiaomi}}
  \fancyfoot[C]{\thepage}
}
\usepackage[most]{tcolorbox}
\tcbset{
  colback=priorlight, colframe=black!15,
  boxrule=0.5pt, arc=6pt,
  left=16pt, right=16pt, top=16pt, bottom=16pt,
  enhanced,
}
\newcommand{\field}[2]{%
  \par\noindent
  {\small\color{black!55}\textsc{#1}}\quad #2\par\addvspace{2pt}}

\newcommand{\ours}{{Xiaomi-TabLDM}\xspace}
\newcommand{\tA}{{TabArena}\xspace}
\newcommand{\tL}{{TALENT}\xspace}

\begin{document}

\thispagestyle{withlogo}  

\begin{center}
{\LARGE\bfseries \ours: A Tabular Foundation Model Technical Report\par}

\vspace{0.8em}
{\large \ours Team}

\end{center}

\vspace{1em}

\begin{tcolorbox}
We introduce \ours, a tabular large data foundation model for classification and regression via in-context learning, which delivers superior prediction accuracy without requiring task-specific fine-tuning. Pretrained exclusively on synthetic data generated from structural causal models (SCMs), our model enables more flexible context utilization and more efficient capacity scaling.

\textbf{i) A new performance standard.}
\textit{Strong regression performance across benchmarks}: \ours ranks 1st on OpenML-CTR23 and 2nd on regression across TALENT, TabArena, and BCCO, demonstrating consistently strong regression performance across four complementary benchmark suites.
\textit{Favorable performance--efficiency trade-off}: \ours combines strong
predictive performance with substantially lower computational cost. For example,
on TabArena regression, it achieves the second-highest Elo while using 82\%
less training time and 68\% less prediction time than the top-ranked TabFM.

\textbf{ii) Large-scale synthetic pretraining.} \ours expands the coverage and diversity of synthetic tabular data used for pretraining. We also adopt a three-stage training strategy together with dual-stream feature grouping, lightweight Attention Residual, and sparse
Mixture-of-Experts, enabling Xiaomi-TabLDM to learn richer feature
interactions and expert specialization across diverse tabular tasks.

\textbf{iii) Test-time scaling.} \ours further extends tabular prediction through test-time compute scaling, where allocating additional computation at inference time consistently improves predictive performance over the base model.

\vspace{0.5em}
\field{Model}{\ours} \field{Date}{September 2026}
\field{GitHub}{https://github.com/xiaomi-research/xiaomi-tabldm }
\field{Hugging Face}{https://huggingface.co/occams/Xiaomi-TabLDM}

\end{tcolorbox}

\begin{figure}[ht!]
    \centering
    \includegraphics[width=\linewidth]{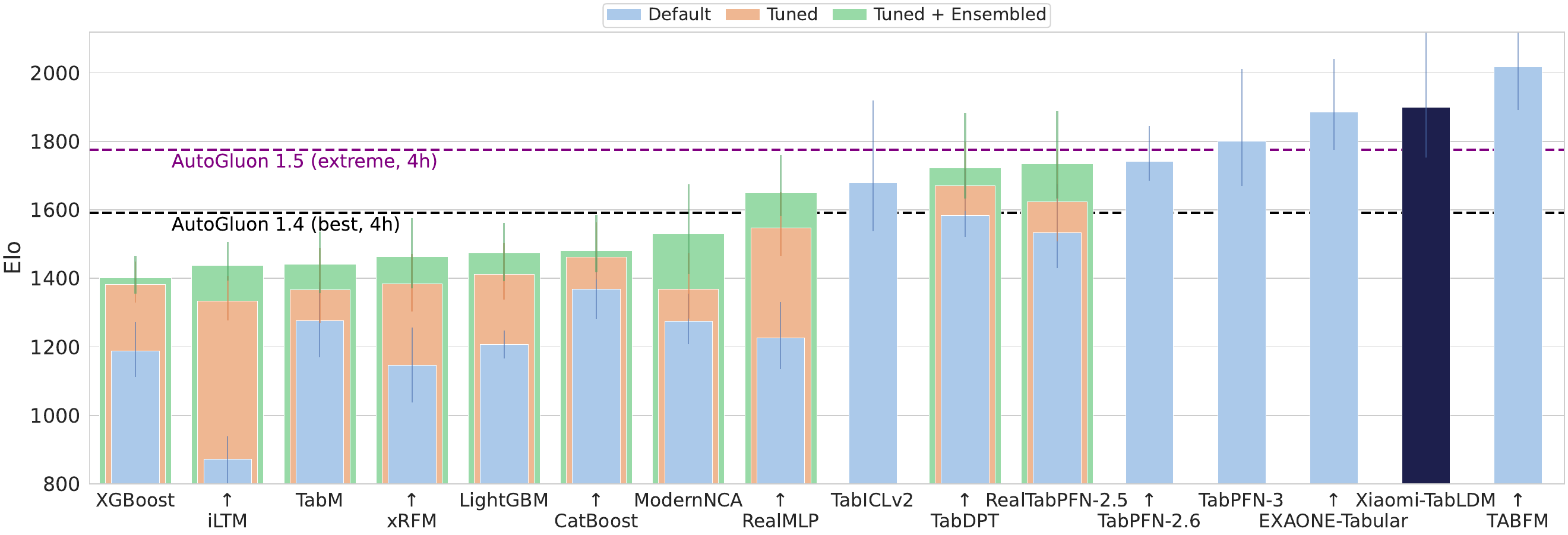}
    \caption{
    Regression Elo performance on \tA (higher is better).
    \ours ranks 2nd among all evaluated methods.
    }
    \label{fig:tA-reg}
\end{figure}

\begin{figure}[ht!]
    \centering
    \includegraphics[width=\linewidth]{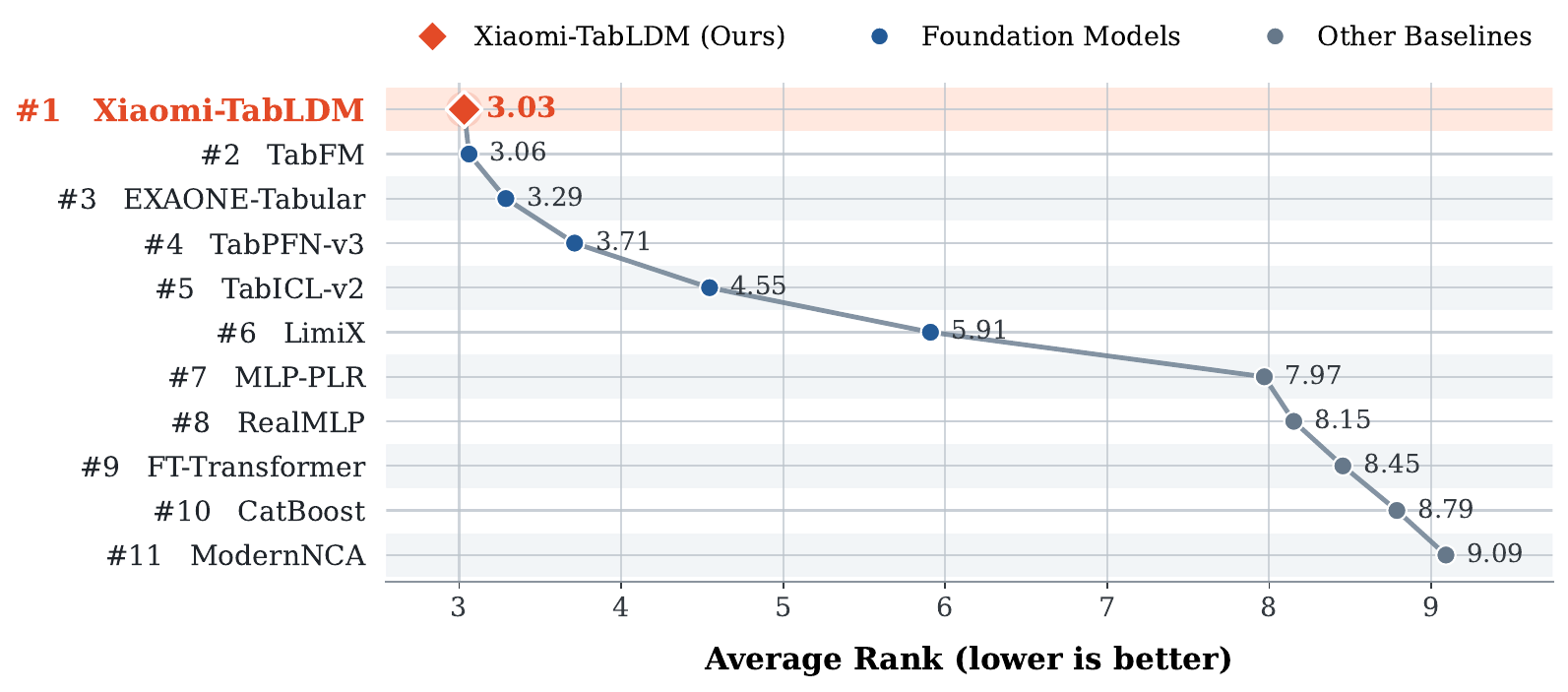}
    \caption{
    Average-rank comparison on OpenML-CTR23 over 33 regression datasets (lower is better).
    \ours achieves the best average rank among all evaluated methods.
    }
    \label{fig:ctr23}
\end{figure}

\begin{figure}[ht!]
    \centering
    \includegraphics[width=\linewidth]{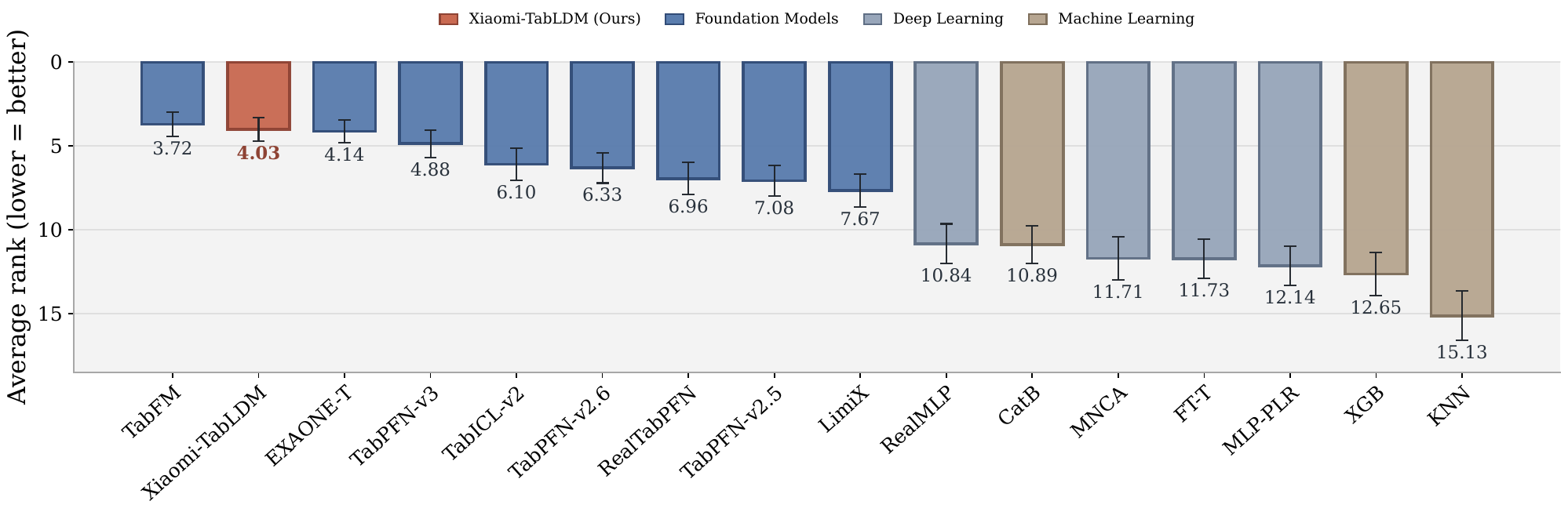}
    \caption{
    Regression average-rank performance on \tL (lower is better).
    \ours ranks 2nd among all evaluated methods.
    }
    \label{fig:tL-reg}
\end{figure}

\section{Introduction}

Structured data is a foundational modality for scientific analysis and operational decision-making across a wide range of domains, including healthcare, finance, logistics, manufacturing, and public policy~\cite{johnson2016mimic,fuster2022predictably,yu2021integrating,krafft2020defining}. Unlike language or perceptual data, tabular data organizes heterogeneous variables under explicit schemas, preserving numerical scales, missingness patterns, categorical structure, and relationships between features that are essential for reliable prediction and quantitative reasoning. As a result, progress on structured-data modeling remains an important and distinct component of the broader development of general-purpose learning systems.

For decades, prediction on tabular data has been dominated by dataset-specific
machine-learning pipelines, particularly gradient-boosted trees and automated
ensembles such as XGBoost~\cite{chen2016xgboost},
LightGBM~\cite{ke2017lightgbm},
CatBoost~\cite{dorogush2018catboost}, and
AutoGluon~\cite{erickson2020autogluon}.
These methods remain strong and reliable, and extensive empirical studies have
shown that boosted trees remain highly competitive with neural networks across
diverse tabular problems~\cite{mcelfresh2023neural}.

Meanwhile, deep learning for tabular data has explored a wide range of
architectures, including attention-based models such as
TabNet~\cite{arik2021tabnet},
AutoInt~\cite{song2019autoint},
TabTransformer~\cite{huang2020tabtransformer}, and
SAINT~\cite{somepalli2022saint};
neural decision and feature-interaction models such as
NODE~\cite{popov2020node},
DANets~\cite{chen2022danets},
DCN-v2~\cite{wang2021dcnv2}, and
T2G-Former~\cite{yan2023t2gformer};
and more recent approaches including
TANGOS~\cite{jeffares2023tangos},
TabCaps~\cite{chen2023tabcaps},
Trompt~\cite{chen2023trompt},
ExcelFormer~\cite{chen2023excelformer},
TabR~\cite{gorishniy2024tabr},
DNNR~\cite{nader2022dnnr}, and
SwitchTab~\cite{wu2024switchtab}.
Despite substantial advances in representation learning, most of these methods
still follow the conventional paradigm of training and tuning a separate model
for each dataset. Recent surveys have further highlighted the growing interest
in adapting large pretrained models and foundation-model paradigms to structured
and tabular data~\cite{fang2024llmtabularsurvey}.
Tabular foundation models~\cite{vanbreugel2024tabular} provide a different
route: a single pretrained model learns across a large distribution of synthetic
or real tabular tasks and directly performs prediction on unseen datasets through
in-context learning, using labeled examples from the target dataset as context
rather than retraining model parameters.

TabPFN established this paradigm by showing that a transformer pretrained over synthetic tabular tasks can perform strong prediction in a single forward pass~\cite{hollmann2022tabpfn}. Subsequent generations have steadily expanded its practical scope. TabPFN v2~\cite{hollmann2025accurate} improved robustness to larger datasets and heterogeneous feature types, while TabPFN-2.5~\cite{grinsztajn2025tabpfn25} further increased the supported number of rows and features and narrowed the gap with heavily tuned machine-learning ensembles. In parallel, TabICL~\cite{qu2025tabicl}, TabDPT~\cite{ma2025tabdpt}, LimiX~\cite{zhang2025limix}, TabFM~\cite{kong2026tabfm}, EXAONE-Tabular~\cite{eo2026exaonetabular} and related approaches have explored larger model capacity, broader training distributions, alternative architectures, and scaling to increasingly diverse tabular settings. Together, these developments have shifted tabular prediction from training a specialized model for each dataset toward building reusable pretrained predictors that transfer across datasets and tasks.

Recent progress in tabular foundation models has increasingly come from jointly scaling training data, model capacity, and inference strategies~\cite{grinsztajn2026tabpfn3technicalreport,qu2026tabiclv2betterfasterscalable}. Following this direction, we explore a broader synthetic prior, more efficient model scaling, and test-time computation within a unified tabular foundation model.
We introduce \ours, a new tabular foundation model that follows and extends this direction. \ours is designed to provide more flexible context utilization while efficiently scaling model capacity. Rather than relying on a single fixed representation pathway, the model learns complementary feature interactions across different granularities and selectively preserves useful information across layers. Sparse expert computation further expands total model capacity while keeping the amount of activated computation limited. Together, these components allow \ours to improve predictive capability without requiring a proportional increase in inference cost.

Across four public benchmark suites, including \tL~\cite{liu2025talent}, 
\tA~\cite{erickson2025tabarena}, BCCO~\cite{zhang2025limix}, 
and OpenML-CTR23~\cite{fischer2023openml}, 
\ours consistently achieves strong performance across diverse classification 
and regression tasks. Its strongest and most consistent results are observed 
on regression: \ours ranks first on OpenML-CTR23 and second on regression 
across \tL, \tA, and BCCO. In particular, on OpenML-CTR23, a benchmark 
dedicated exclusively to tabular regression, \ours achieves the best average 
rank among all evaluated methods. On \tA, \ours achieves the second-highest 
regression Elo, outperforming strong recent tabular foundation models including 
TabPFN-3, TabPFN-2.6, and TabICL-v2.

Beyond regression, \ours also remains highly competitive across mixed-task 
evaluations. On \tL, it ranks second overall and achieves the best performance 
on binary classification. On \tA, it ranks fourth overall across 51 datasets 
and 816 tasks, while on BCCO it maintains top-tier aggregate performance across 
classification and regression settings.

Beyond predictive performance, \ours provides a favorable 
performance--efficiency trade-off. On TabArena regression, \ours achieves the 
second-highest Elo while requiring 82\% less training time and 68\% less 
prediction time than TabFM. At a model scale comparable to TabPFN-3, \ours 
also achieves stronger predictive performance, with particularly pronounced 
gains on regression. These results show that \ours combines 
consistently strong regression performance with competitive overall performance 
and efficient model scaling.

\section{Overall Framework}

\ours takes a tabular dataset as input and predicts test targets by conditioning on labeled training examples as in-context demonstrations, without task-specific training.  Its design centers on three components: large-scale synthetic pretraining, an efficient architecture that combines flexible context utilization with sparse expert specialization, and test-time compute scaling.  

\subsection{Architecture}

\begin{figure}[t!]
  \centering
    \includegraphics[width=0.98\linewidth]{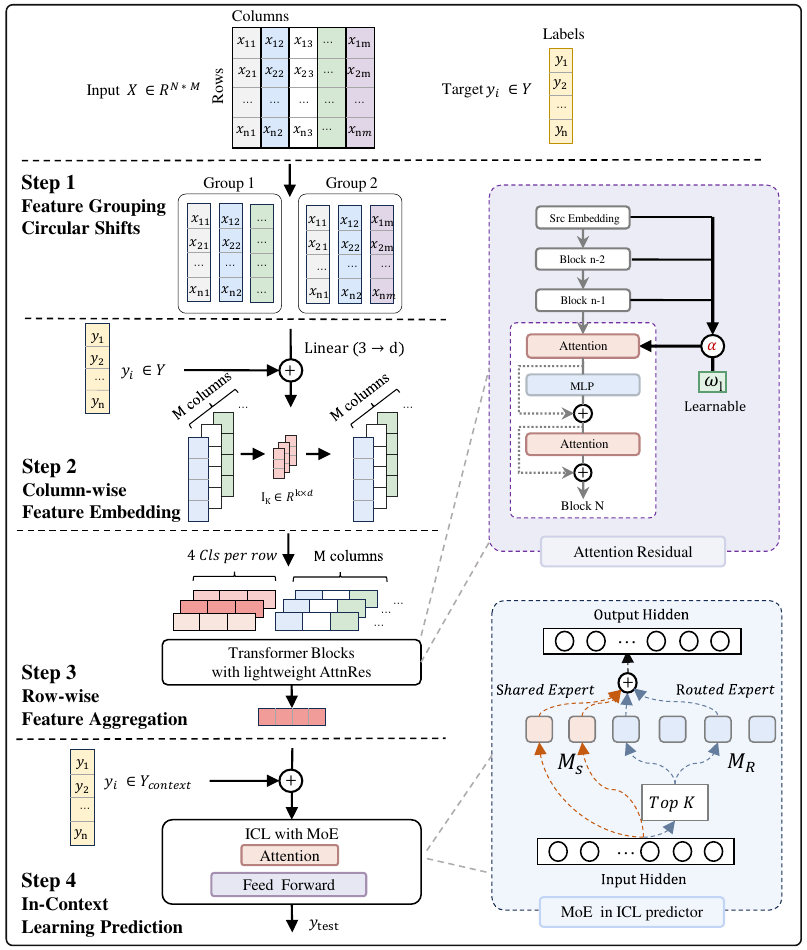}
  \caption{Overall architecture of \ours, which introduces an ICL architecture with a dual-stream feature grouping scheme, lightweight Attention Residual connections, and a sparse Mixture-of-Experts (MoE) design, where a router activates the top-$K$ of $M_R$ routed experts per token together with $M_S$ shared experts.}
  \label{fig:onetab}
  \vspace{0.8em}
\end{figure}

The full architecture of \ours is shown in Figure~\ref{fig:onetab}. \ours is a transformer-based~\cite{vaswani2017attention} in-context learning (ICL) model that processes a table through three sequential stages: column-wise feature embedding, row-wise aggregation, and in-context learning prediction.


 \begin{itemize}[leftmargin=1.5em]
      \item \textbf{Column-wise feature embedding.} We first organize the input
      features into a dual-stream feature grouping,  which breaks feature symmetries and yields column representations that are more flexible and robust to feature permutation.
      Each grouped column is then encoded independently by a Set
      Transformer~\cite{lee2019set}, whose inducing-point attention summarizes it
      through a small set of learned inducing tokens instead of attending across
      all rows, producing the final column embeddings.

      \item \textbf{Row-wise feature aggregation.} For each row, the per-column
      embeddings are prepended with a set of learnable CLS tokens and processed by
      a stack of self-attention layers. Within these layers, we introduce lightweight Attention Residual (AttnRes) connections in place of the plain additive residual, where the outputs of the preceding residual blocks are adaptively fused to improve gradient flow and representation reuse across depth. Finally, the CLS tokens query the full sequence, and their concatenated states form one fixed-length vector per row.

      \item \textbf{In-context learning prediction.} The row embeddings for the
      training and test sets are jointly passed to a transformer that performs
      ICL: training rows attend to one another to model intra-set structure, while
      test rows attend only to the training rows to form their predictions. The
      same lightweight AttnRes connections are retained here, and the
      feed-forward sub-layer of selected layers is replaced with a sparse
      Mixture-of-Experts, strengthening the model's ability to adapt to
      heterogeneous datasets.
  \end{itemize}

In the Column-wise feature embedding and In-context learning prediction stages, attention uses the query-aware scalable softmax (QASSMax)\cite{qu2026tabiclv2betterfasterscalable}, which rescales queries by input length to improve length generalization to large training sets. Further stage-specific technical details of \ours's design are presented below:

\begin{itemize}[leftmargin=1.5em]

\item\textbf{Dual-stream feature grouping.}  
\ours groups each column with two other columns selected by circular shifts over the $m$ feature columns and encodes the triple with a shared linear layer $\mathrm{Linear}(\mathbb{R}^{3}\to\mathbb{R}^{d})$. Given an offset set $\mathcal{S}=(\delta_1,\delta_2,\delta_3)$, the embedding of column $j$ is 
 \begin{equation}
 E[i,j]=\operatorname{Linear}\big(x_{i,(j+\delta_1)\bmod m},x_{i,(j+\delta_2)\bmod m},x_{i,(j+\delta_3)\bmod m}\big)
  \end{equation}
 \ours runs two such streams that differ only in  $\mathcal{S}$. The first stream follows TabICLv2\cite{qu2026tabiclv2betterfasterscalable} with fixed dyadic offsets $(1,2,4)$, while the second uses width-adaptive offsets spread geometrically toward a target span $s$, $(1,\lfloor\sqrt{s}\rfloor,s)$ with $s=\min(s_{\max},m-1)$, so it stretches as wide as the table allows without wrapping past it. When the table has few columns, distinct offsets can map to the same column modulo $m$. Such collisions are resolved by shifting one offset so that the three grouped columns stay distinct and the two streams remain complementary. The resulting embeddings from the two streams are then summed to form the column-wise input embedding.

\item\textbf{Lightweight AttnRes connections.} We introduce a lightweight inter-layer connection that improves how information propagates across depth, inspired by Block AttnRes~\cite{team2026attention}. A standard residual connection feeds each layer only the previous output, aggregating all earlier outputs with fixed, uniform weights. The lightweight AttnRes replaces this fixed sum with learned, input-dependent weights, letting each layer selectively retrieve the earlier representations it needs. For efficiency, the layers are partitioned into blocks and the  mechanism is applied only at a fixed stride in depth, so a layer attends over the completed block representations rather than every prior layer. At layer $l$, a learnable pseudo-query $w_l$ scores each candidate block, and a softmax over depth turns these scores into the aggregation weights:
 \begin{equation}
  \alpha^{l}_{j}=\frac{\exp\left(w_l^{\top}\mathrm{RMSNorm}(b_j)\right)}{\sum_{i=0}^{l-1}\exp\left(w_l^{\top}\mathrm{RMSNorm}(b_i)\right)}
  \end{equation}
This gives selected layers learned access to both shallow and deep features, improving gradient flow and stabilizing training.

\item\textbf{ Sparse Mixture-of-Experts.} We redesign the ICL predictor by replacing the feed-forward network of selected layers with a sparse Mixture-of-Experts (MoE)\cite{dai2024deepseekmoe,lepikhin2020gshard} that scales capacity for greater representational flexibility. For each row token, a linear router scores the $M_R$ routed experts and a top-$K$ operator activates the $K$ highest-scoring ones, whose outputs are combined with those of $M_S$ shared experts applied to every token. Since only $K$ experts are activated per token, the parameter count scales with $M_R$ while the per-token cost stays proportional to $K$ feed-forward layers, independent of $M_R$. The shared experts learn the common transformation reused across all tokens, so that the routed experts, rather than redundantly relearning it, can devote their capacity to specialization.

\end{itemize}

\subsection{Pretraining}
We pretrain classification and regression models separately on synthetically generated tabular datasets, using a three-stage curriculum that progressively scales the dataset size. The classifier is trained with a cross-entropy loss and the regressor with a pinball loss over $999$ quantiles. 
\begin{itemize}[leftmargin=*]
        \item \textbf{Stage 1.} $500\text{K}$ steps for classification and $300\text{K}$ for regression, on datasets of $1{,}024$ samples with $30\text{--}90\%$ used for training, a maximum learning rate of $8\text{e-}4$, and gradient clipping at $10$. This stage uses the default feed-forward blocks. Enabling AttnRes stabilizes convergence and accounts for the shorter regression schedule; without it, the model collapses at an earlier step on large datasets (roughly beyond $1{,}200$ samples).
        \item \textbf{Stage 2.} $40\text{K}$ steps on datasets of $400\text{--}10{,}240$
        samples (log-uniform), $\sim\!80\%$ used for training, a maximum learning
        rate of $1\text{e-}4$, and gradient clipping at $10$. MoE and its auxiliary
        losses are enabled from this stage on.
        \item \textbf{Stage 3.}  $10\text{K}$ steps on datasets of $400\text{--}60{,}000$
        samples (log-uniform), $\sim\!80\%$ used for training, a maximum learning
        rate of $2\text{e-}5$, and gradient clipping at $1$, adapting the
        model to long-context, large-sample inference.
  \end{itemize}
 \paragraph{Optimizer and pre-training cost.}We train with the Muon optimizer\cite{jordan2024muon} based on the implementation of TabICLv2\cite{qu2026tabiclv2betterfasterscalable}, together with a cosine learning-rate schedule and a weight decay of $0.01$. Pre-training runs on 8 A100 GPUs (80\,GB): around 7--10 days for \textbf{Stage 1}, and 2 days each for \textbf{Stage 2} and \textbf{Stage 3}, with FlashAttention-2\cite{dao2024flashattention} enabled from \textbf{Stage 2} onward.

\subsection{Synthetic Prior}
  \label{sec:synthetic_prior}

\begin{figure}[!ht]
    \centering
    \includegraphics[width=1.0\linewidth]
    {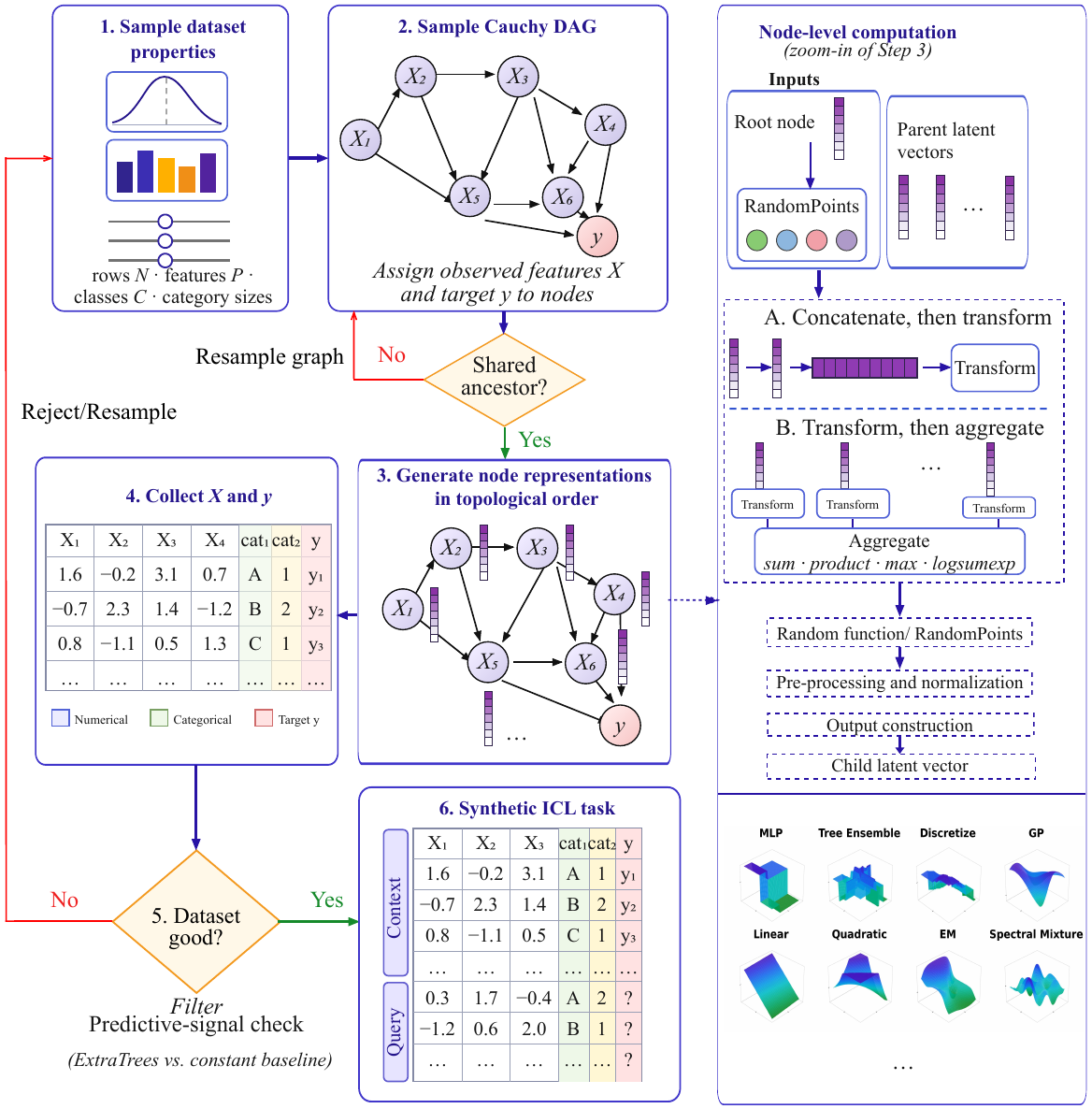}
    \caption{\textbf{Schematic overview of the synthetic data generation prior.}
The dataset-level pipeline samples dataset properties and a Cauchy DAG, checks
whether the selected features and target share an ancestor, generates node
representations in topological order, and collects the resulting tabular data.
Datasets passing the predictive-signal check are split into context and query
sets to form synthetic in-context learning tasks. The right panel details the
node-level computation and the 12 random-function classes used by the prior.}
    \label{fig:synthetic_prior}
\end{figure}

  Following previous tabular foundation models
~\cite{hollmann2022tabpfn,qu2025tabicl},
\ours is pretrained entirely on synthetic datasets generated from a
structural causal model (SCM) prior
~\cite{pearl2009causality,peters2017elements}.
The prior is designed to maximize the diversity of dataset structures,
variable types, and functional relationships while retaining learnable
predictive signals~\cite{zhang2025mitra}. Figure~\ref{fig:synthetic_prior} summarizes the complete generation pipeline.

We organize its main components into the following four stages:

\begin{enumerate}[leftmargin=*]
    \item \textbf{Dataset configuration.}
    We first sample the global properties of each task, including its size,
    numerical and categorical feature counts, categorical cardinalities, target
    type, and context--query split.

    \item \textbf{Graph-based data generation.}
We sample a directed acyclic graph using the random Cauchy graph mechanism
of TabICL-v2 \citep{qu2026tabiclv2betterfasterscalable}. Latent
representations are propagated through the graph in topological order using
the node computations illustrated in the middle row of
Figure~\ref{fig:synthetic_prior}, with additional Gaussian noise injected to
increase data diversity and robustness. We also add additional transformation functions to broaden the range of
smoothness, sparsity, additivity, and local structure represented by the prior.

    \item \textbf{Tabular data construction.}
    We assign the observed features and target to sampled graph nodes and collect
    their values after evaluating the graph. Numerical features and regression
    targets are obtained from continuous latent dimensions, whereas categorical
    features and classification targets are produced through discretization or
    categorical sampling. As illustrated in Figure~\ref{fig:synthetic_prior},
    only selected dimensions are observed. The remaining dimensions introduce
    latent variation. Random scaling further varies the relative importance of
    features and graph nodes across tasks.

    \item \textbf{Postprocessing and quality filtering.}
    We standardize and randomly permute the resulting tables before removing
    invalid or uninformative tasks. A graph is resampled when the features and
    target share no common ancestor. We also reject tasks for which an ExtraTrees
    model \citep{geurts2006extremely} cannot reliably outperform a constant
    predictor. Finally, we divide each retained dataset into context and query
    samples for in-context pretraining.
\end{enumerate}

Together, these stages generate diverse classification and regression tasks by
varying dataset configurations, dependency structures, functional relationships,
and observed feature types.

\section{Test-Time Scaling}

Test-time scaling improves predictive performance by allocating additional
inference-time computation while keeping the pretrained model fixed. Recent
tabular foundation models have implemented this idea by aggregating predictions
across multiple estimators, dataset permutations, and feature transformations
\citep{grinsztajn2026tabpfn3technicalreport,qu2026tabiclv2betterfasterscalable}.
TabFM further combines predictions generated from SVD and cross-feature
representations using nonnegative least squares \citep{kong2026tabfm}.
Building on these approaches, we introduce a context-adaptive framework that
enhances inference based on the given context. This allows the inference pipeline to adapt to heterogeneous
datasets without updating the pretrained model.

Our test-time scaling strategy consists of the following components:

\begin{enumerate}[leftmargin=*]
    \item \textbf{Feature shuffling and sampling.}
    Each ensemble member uses a randomly shuffled feature order and randomly samples a subset of features for training and inference. This reduces sensitivity to column ordering and specific feature combinations, creates prediction diversity at negligible additional cost, and enhances robustness in high-dimensional scenarios.

    \item \textbf{Diverse preprocessing views.}
    We construct complementary views through different normalization schemes,
    quantile transformation, SVD-based representations, and feature
    interactions. For regression, we additionally consider target
    transformations for heavy-tailed outcomes.

    \item \textbf{NNLS-based combination.}
    For both classification and regression, we learn nonnegative weights for the
estimators corresponding to the retained inference modes.  For large datasets, we use predictions on this single
    holdout set to estimate the weights. Otherwise, we generate out-of-fold
    predictions through \(K\)-fold cross-validation to obtain a larger and more
    stable sample for weight estimation. We then fit nonnegative least squares
    to one-hot labels for classification or continuous targets for regression
    and use the resulting weights to combine the candidate inference views.

    \item \textbf{Final classification adjustment.}
    The combined class probabilities can be calibrated on validation data to
    improve probabilistic prediction quality. Final class labels are then
    obtained by taking the class with the largest adjusted probability.
\end{enumerate}

Overall, the procedure scales inference by increasing the diversity and number of evaluated prediction paths, rather than by modifying the underlying model. The NNLS-based combination step effectively exploits complementary predictions among the diverse inference views, allowing the method to flexibly adapt to heterogeneous datasets. Thus, the approach uses additional test-time computation selectively, harnessing the benefits of ensemble diversity while mitigating the risk of overfitting through the nonnegative constraint on combination weights.

\section{Evaluation}

In this section, we evaluate \ours on four public tabular learning benchmarks, 
\tL~\cite{liu2025talent}, \tA~\cite{erickson2025tabarena}, BCCO~\cite{zhang2025limix}, 
and OpenML-CTR23~\cite{fischer2023openml}, covering 
a broad range of real-world classification and regression tasks. 




Table~\ref{tab:db} summarizes the benchmark collections used in our evaluation, covering four benchmark suites with complementary task types and dataset scales. TALENT is the largest collection, containing 300 datasets, including 200 classification and 100 regression tasks, while BCCO contains 156 datasets with a similar mixture of classification and regression problems. TabArena provides 51 datasets spanning both task types. In addition, OpenML-CTR23 focuses exclusively on regression. Across these benchmarks, most datasets contain fewer than 10K training examples, with a smaller portion ranging from 10K to 100K; TALENT additionally includes several datasets with more than 100K training instances.


In Section~\ref{sec:tL}, we evaluate \ours on TALENT, reporting regression,
overall, and binary-classification performance. In Section~\ref{sec:tA}, we
report the overall and regression results on TabArena and further
analyze performance--efficiency trade-offs and pairwise model
comparisons. We further evaluate \ours on BCCO and OpenML-CTR23
in Sections~4.3 and~4.4, respectively.

\begin{figure}[ht!]
    \centering

    \includegraphics[width=\linewidth]{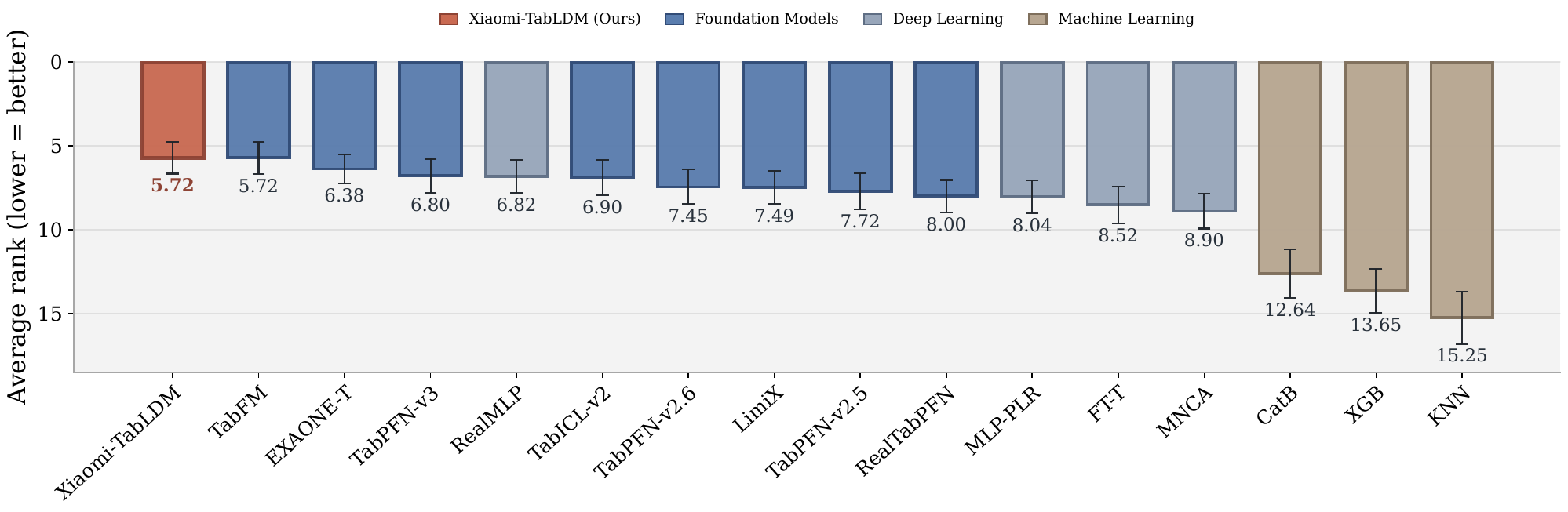}

    \vspace{1.0em}

    \includegraphics[width=\linewidth]{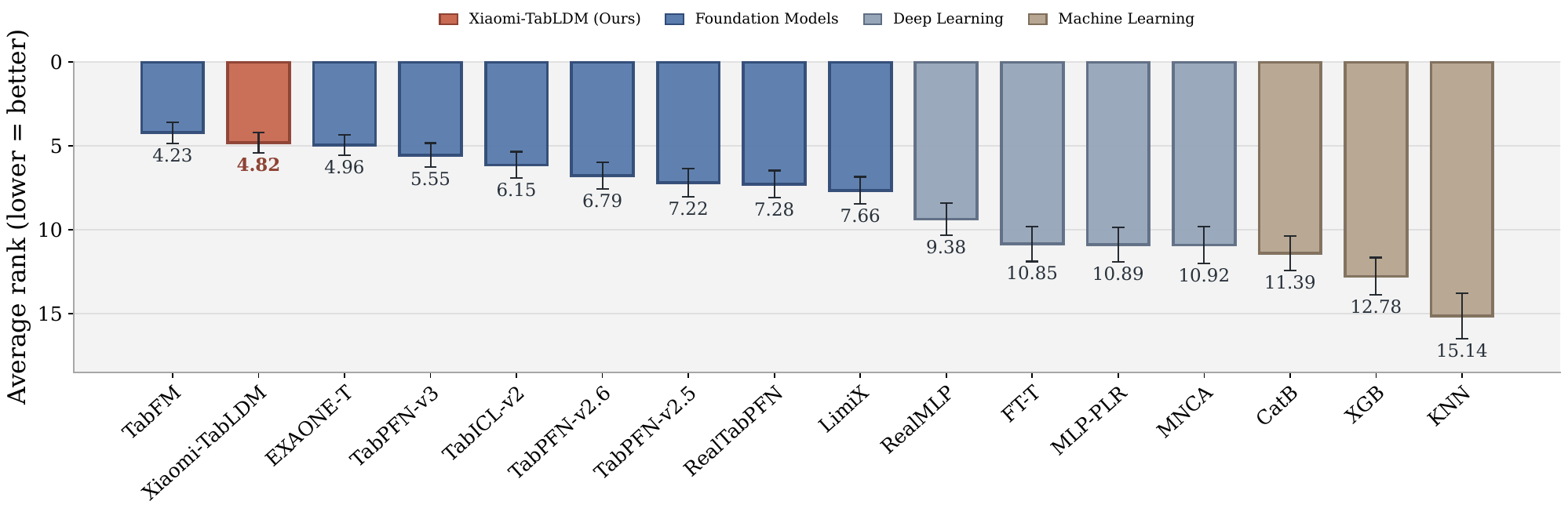}

    \caption{
    Performance on the TALENT benchmark following the TabICLv2 evaluation
    protocol~\cite{qu2026tabiclv2betterfasterscalable}.
For methods with missing results on a subset of datasets, the corresponding
score entries are imputed using K-nearest-neighbour values following the
evaluation protocol.
    The top panel reports binary-classification performance, while the bottom
    panel reports overall performance across task types.
    Results are measured by average rank across datasets (lower is better);
    error bars indicate 95\% bootstrap confidence intervals.
    }
    \label{fig:talent_results}
\end{figure}

\begin{table}[htbp]
    \centering
    \caption{Statistics of the benchmark suites considered in our evaluation.
We report the total number of datasets, task-type composition, classification-task breakdown, and dataset-size distribution for each benchmark collection.}
    \label{tab:dataset_statistics}
    \resizebox{\linewidth}{!}{
    \begin{tabular}{lcccccccc}
        \toprule
        \multirow{2}{*}{Benchmark}
        & \multirow{2}{*}{\# Datasets}
        & \multicolumn{2}{c}{Task Type}
        & \multicolumn{2}{c}{Classification Type}
        & \multicolumn{3}{c}{Training Set Size} \\
        \cmidrule(lr){3-4}
        \cmidrule(lr){5-6}
        \cmidrule(lr){7-9}
        & & Cls. & Reg.
        & Binary & Multiclass
        & $<10$K & 10K--100K & $>100$K \\
        \midrule
        TALENT       & 300 & 200 & 100 & 120 & 80 & 218 & 78 & 4 \\
        BCCO         & 156 & 106 &  50 &  71 & 35 & 128 & 28 & - \\
        TabArena     &  51 &  38 &  13 &  30 &  8 &  36 & 15 & - \\
        OpenML-CTR23 &  33 &   - &  33 &   - &  - &  23 & 10 & - \\
        \bottomrule
    \end{tabular}
    }
    \label{tab:db}
\end{table}

\subsection{\tL}~\label{sec:tL}
TALENT~\cite{liu2025talent} is a unified benchmark and toolbox for evaluating tabular learning methods under consistent preprocessing, hyperparameter tuning, and evaluation protocols. Its benchmark suite contains 300 datasets spanning 120 binary classification, 80 multiclass classification, and 100 regression tasks, covering diverse dataset sizes and application domains. TALENT includes a broad range of classical machine-learning, tree-based, deep tabular, and recent foundation-model approaches, providing a complementary evaluation setting to TabArena for assessing model performance across different types of tabular prediction tasks.



\subsubsection{Main Results}

\paragraph{Regression performance.}
\ours performs particularly strongly on regression, achieving an average rank
of \textbf{4.03} across 100 regression datasets and ranking \textbf{second}
among all evaluated methods, behind only TabFM (3.72). It outperforms other
strong tabular foundation models, including EXAONE-Tabular (4.15),
TabPFN-v3 (4.88), TabICL-v2 (6.10), and TabPFN-v2.6 (6.33).
Together with its strong results on other regression benchmarks, this
demonstrates the consistent regression capability of \ours across diverse
tabular datasets.

\paragraph{Overall performance.}
As shown in Figure~\ref{fig:talent_results} (bottom), \ours achieves an
average rank of \textbf{4.82}, ranking \textbf{second overall}, behind only
TabFM (4.23). It outperforms other recent tabular foundation models, including
EXAONE-Tabular (4.96), TabPFN-v3 (5.55), TabICL-v2 (6.15), and
TabPFN-v2.6 (6.79). These results show that the strong regression performance
of \ours extends to competitive performance across the broader range of
tabular tasks covered by TALENT.

\paragraph{Classification.}
As shown in Figure~\ref{fig:talent_results} (top), \ours also performs
strongly on binary classification, achieving an average rank of \textbf{5.72}
and tying with TabFM for the best performance among all evaluated methods.
It further outperforms EXAONE-Tabular (6.38), TabPFN-v3 (6.80), and
TabICL-v2 (6.90).



\begin{figure}[t!]
    \centering
    \begin{subfigure}[b]{0.48\linewidth}
        \centering
        \includegraphics[width=1\linewidth]{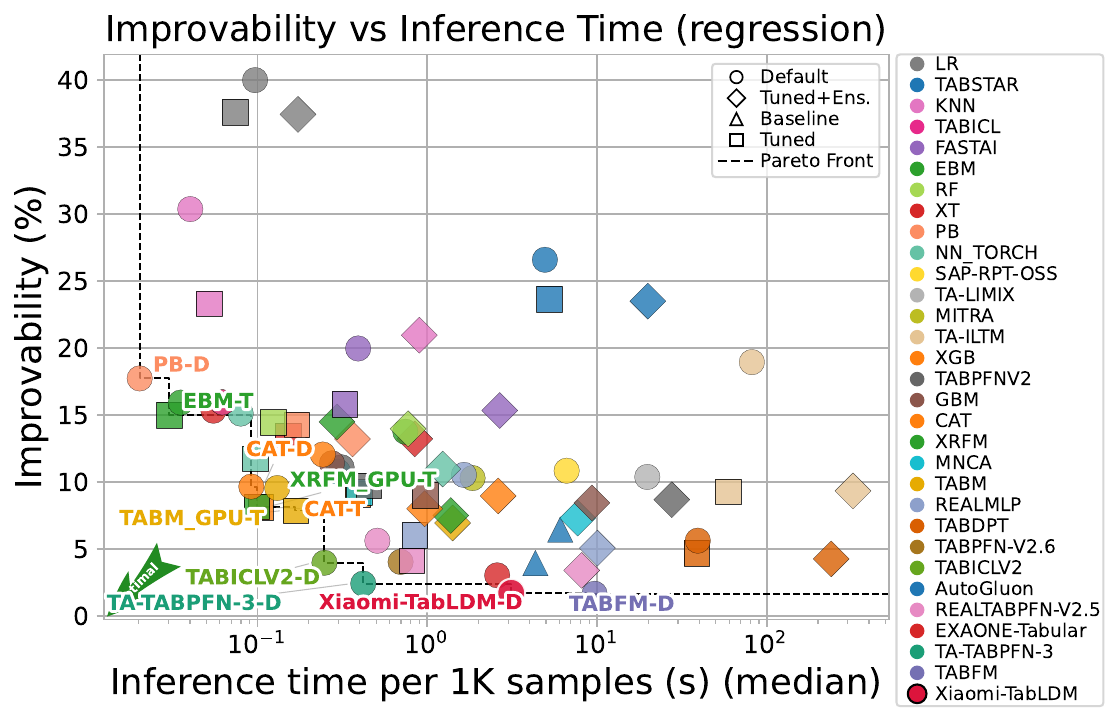}
        \caption{Inference-efficiency and improvability trade-offs on regression tasks.
        Improvability measures how much worse a model is than the best per-dataset model. 
        See \S~\ref{sec:tA} for more details.}
        \label{fig:tradeoff-regression}
    \end{subfigure}
    \hfill
    \begin{subfigure}[b]{0.48\linewidth}
        \centering
        \includegraphics[width=1\linewidth]{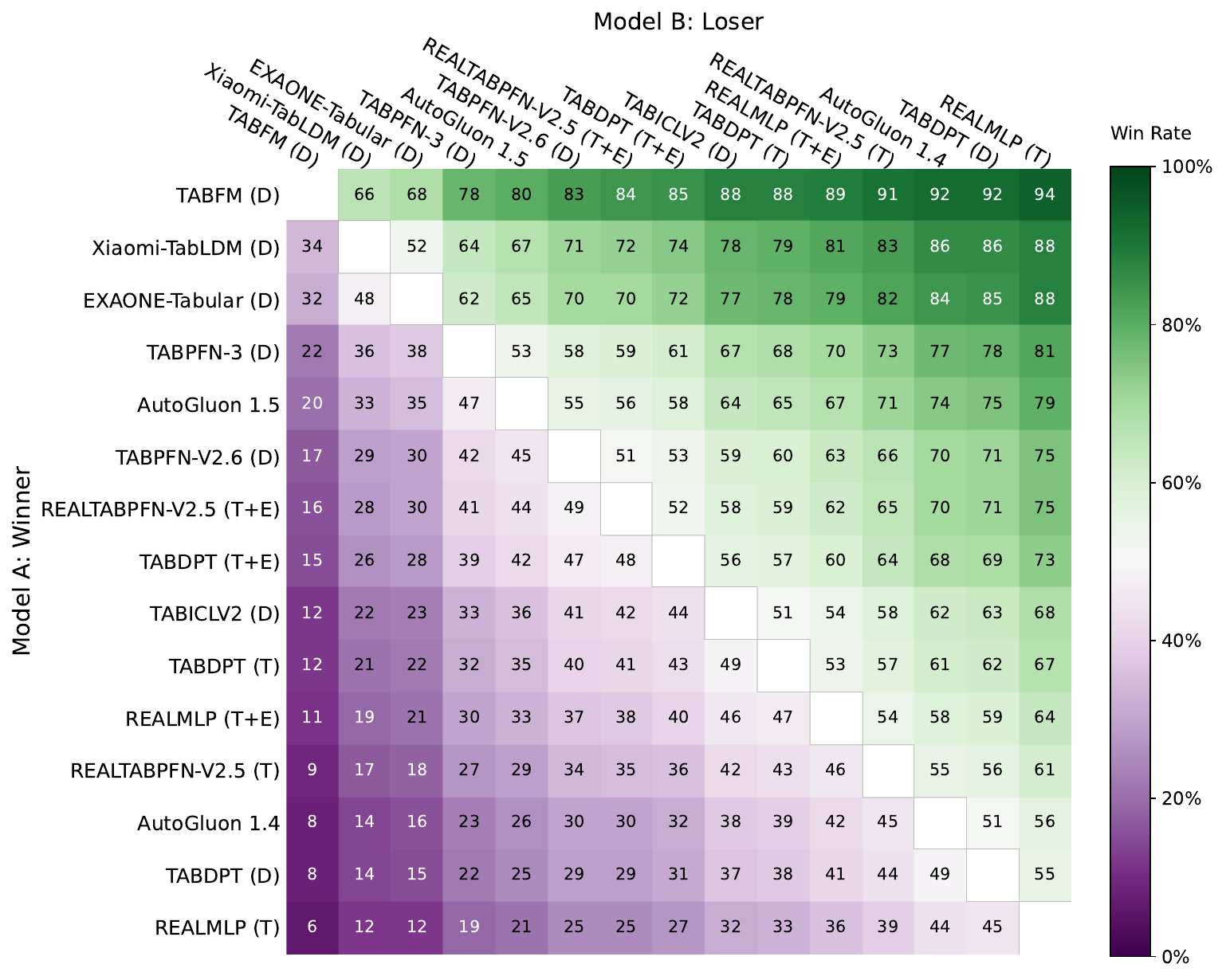}
\caption{
Pairwise win-rate matrix on the TabArena regression subset.
Each entry $(i,j)$ reports the proportion of regression instances on which
model $i$ outperforms model $j$. Diagonal entries denote self-comparison
(50\% by definition).
}
        \label{fig:winrate-regression}
    \end{subfigure}
    
    \caption{Regression task performance analysis: (left) efficiency-improvability trade-offs; (right) pairwise model comparisons.}
    \label{fig:tradeoff-pareto}
\end{figure}

\begin{table}[t!]
\centering
\caption{
Performance on the regression subset of TabArena, covering 13 datasets.
We compare models in terms of Elo, number of wins, improvability, and training and prediction time. Here \textbf{D} denotes the default (untuned) model, \textbf{T} the fine-tuned model, and \textbf{T+E} ensembling after fine-tuning. \ours achieves the second-highest Elo while maintaining substantially lower computational cost than most tuned and ensembled baselines.
}
\label{tab:tabarena_reg}
\small
\setlength{\tabcolsep}{6pt}
\renewcommand{\arraystretch}{1.05}

\begin{tabular}{llccrr}
\toprule
\textbf{Model} & \textbf{Elo ($\uparrow$)} & \textbf{\#wins ($\uparrow$)} & \textbf{Improva-} & \textbf{Train time} & \textbf{Predict time} \\
 &  &  & \textbf{bility ($\downarrow$)} & \textbf{per 1K [s]} & \textbf{per 1K [s]} \\
\midrule

TABFM (D)
& \textcolor{gold}{\textbf{2019${}_{-127,+181}$}}
& \textcolor{gold}{\textbf{3.7}}
& \textcolor{gold}{\textbf{1.6\%}}
& 38.85
& 9.67 \\

\textbf{\ours (D)}
& \textcolor{silver}{\textbf{1900${}_{-148,+278}$}}
& \textcolor{silver}{\textbf{1.8}}
& \textcolor{silver}{\textbf{1.7\%}}
& 6.99
& 3.12 \\

EXAONE-Tabular (D)
& \textcolor{bronze}{\textbf{1885${}_{-110,+155}$}}
& \textcolor{bronze}{\textbf{1.5}}
& 3.0\%
& 13.71
& 2.58 \\

TabPFN-3 (D)
& 1800${}_{-131,+211}$
& 0.9
& \textcolor{bronze}{\textbf{2.4\%}}
& 3.87
& 0.42 \\

AutoGluon 1.5 (extreme, 4h)
& 1776${}_{-96,+137}$
& 1.3
& 3.9\%
& 335.03
& 4.33 \\

TabPFN-2.6 (D)
& 1741${}_{-56,+104}$
& 0.1
& 4.0\%
& 8.52
& 0.70 \\

RealTabPFN-2.5 (T+E)
& 1736${}_{-103,+153}$
& 0.2
& 3.4\%
& 1709.05
& 8.12 \\

TabDPT (T+E)
& 1722${}_{-90,+160}$
& \textcolor{bronze}{\textbf{1.5}}
& 4.3\%
& 4786.60
& 239.30 \\

TabICLv2 (D)
& 1679${}_{-142,+242}$
& 0.5
& 4.0\%
& 2.10
& 0.25 \\

TabDPT (T)
& 1670${}_{-73,+128}$
& 0.0
& 4.7\%
& 4786.60
& 38.50 \\

RealMLP (T+E)
& 1650${}_{-67,+111}$
& 0.1
& 5.1\%
& 3995.01
& 10.05 \\

RealTabPFN-2.5 (T)
& 1624${}_{-117,+154}$
& 0.2
& 4.1\%
& 1709.05
& 0.81 \\

AutoGluon 1.4 (best, 4h)
& 1592${}_{-90,+117}$
& 0.0
& 6.4\%
& 1866.35
& 6.07 \\

TabDPT (D)
& 1584${}_{-64,+137}$
& 0.0
& 5.6\%
& 46.62
& 39.21 \\

RealMLP (T)
& 1547${}_{-82,+105}$
& 0.0
& 6.0\%
& 3995.01
& 0.84 \\

RealTabPFN-2.5 (D)
& 1534${}_{-106,+141}$
& 0.0
& 5.6\%
& 7.04
& 0.51 \\

ModernNCA (T+E)
& 1531${}_{-118,+145}$
& 0.6
& 7.3\%
& 3779.70
& 7.69 \\

CatBoost (T+E)
& 1482${}_{-65,+103}$
& 0.0
& 8.0\%
& 3555.27
& 0.96 \\

LightGBM (T+E)
& 1474${}_{-83,+89}$
& 0.0
& 8.4\%
& 700.19
& 9.32 \\

xRFM (T+E)
& 1464${}_{-95,+112}$
& 0.0
& 7.5\%
& 714.50
& 1.38 \\

TabM (T+E)
& 1442${}_{-85,+130}$
& 0.0
& 6.9\%
& 4160.58
& 1.41 \\

XGBoost (T+E)
& 1402${}_{-47,+64}$
& 0.0
& 9.0\%
& 834.93
& 2.61 \\

CatBoost (D)
& 1369${}_{-89,+93}$
& 0.0
& 9.7\%
& 10.89
& 0.09 \\

ModernNCA (D)
& 1274${}_{-68,+81}$
& 0.0
& 10.8\%
& 15.50
& 0.30 \\

LimiX (D)
& 1246${}_{-154,+167}$
& 0.1
& 10.4\%
& 74.68
& 19.76 \\

RealMLP (D)
& 1226${}_{-91,+107}$
& 0.0
& 10.5\%
& 8.90
& 1.64 \\

LightGBM (D)
& 1206${}_{-41,+41}$
& 0.0
& 11.4\%
& 2.11
& 0.27 \\

XGBoost (D)
& 1189${}_{-77,+83}$
& 0.0
& 12.0\%
& 2.24
& 0.24 \\
TabPFNv2 (D)
& 1184${}_{-143,+133}$
& 0.0
& 11.1\%
& 2.80
& 0.31 \\

RandomForest (T+E)
& 1162${}_{-61,+62}$
& 0.0
& 14.0\%
& 515.75
& 0.77 \\

xRFM (D)
& 1147${}_{-109,+110}$
& 0.0
& 13.8\%
& 2.45
& 0.74 \\

ExtraTrees (D)
& 1069${}_{-107,+94}$
& 0.0
& 15.3\%
& 0.47
& 0.06 \\

RandomForest (D)
& 1000${}_{-71,+44}$
& 0.0
& 15.9\%
& 0.53
& 0.06 \\

FastaiMLP (D)
& 864${}_{-160,+111}$
& 0.0
& 20.0\%
& 2.60
& 0.39 \\

KNN (D)
& 680${}_{-246,+170}$
& 0.0
& 30.4\%
& 0.19
& 0.04 \\

Linear (D)
& 295${}_{-413,+146}$
& 0.0
& 40.0\%
& 0.95
& 0.10 \\
\bottomrule
\end{tabular}
\end{table}
\label{tab:reg}



\subsection{TabArena}~\label{sec:tA}

We also evaluate \ours on \tA~\cite{erickson2025tabarena}, consisting of
51 datasets and 816 complete tasks, including 38 classification datasets and
13 regression datasets. We use Elo as the primary evaluation metric and
additionally report the number of dataset wins, improvability, and training
and prediction time. We report both overall and regression performance, and
further analyze performance--efficiency trade-offs and pairwise model
comparisons.

Our primary comparisons include leading gradient-boosted tree frameworks,
including XGBoost~\cite{chen2016xgboost},
CatBoost~\cite{dorogush2018catboost}, and
LightGBM~\cite{ke2017lightgbm}, as well as strong recent tabular foundation
models, including TabICLv2~\cite{qu2026tabiclv2betterfasterscalable},
TabPFN~\cite{hollmann2022tabpfn},
TabPFN-3~\cite{grinsztajn2026tabpfn3technicalreport},
RealTabPFN~\cite{garg2025realtabpfn}, and
TabFM~\cite{kong2026tabfm}. We also compare against
AutoGluon~\cite{erickson2020autogluon} as a strong AutoML baseline.

\subsubsection{Main Results}

\paragraph{Regression.}
On the 13 regression datasets, as shown in Table~\ref{tab:tabarena_reg},
\ours achieves an Elo score of \textbf{1900}, ranking \textbf{second}
among all evaluated methods by Elo point estimate, behind only TabFM (2019).
It outperforms other strong tabular foundation models, including
EXAONE-Tabular (1885), TabPFN-3 (1800), TabPFN-2.6 (1741), and
TabICLv2 (1679), as well as the heavily tuned AutoGluon 1.5 extreme
configuration (1776).

Beyond Elo, \ours achieves \textbf{1.8 dataset wins} and the
\textbf{second-lowest improvability} of \textbf{1.7\%}, only marginally
behind TabFM (1.6\%). The pairwise win-rate matrix in
Figure~\ref{fig:tradeoff-pareto}(b) provides a complementary view:
\ours wins \textbf{67\%} of pairwise comparisons against AutoGluon 1.5
and \textbf{78\%} against TabICLv2. Together, these results demonstrate
strong and consistent regression performance across the TabArena benchmark.

\paragraph{Overall Performance.}
Across all 51 datasets and 816 tasks, as shown in
Table~\ref{tab:elo_overall_all}, \ours achieves an Elo score of
\textbf{1659}, ranking \textbf{fourth} by Elo point estimate.
Its performance is nearly tied with AutoGluon 1.5 extreme (1662) and
surpasses TabPFN-3 (1650), TabPFN-2.6 (1596),
RealTabPFN-2.5 with tuning and ensembling (1579), and TabICLv2 (1576).
Moreover, \ours achieves \textbf{3.1 dataset wins}, exceeding TabPFN-3
and other recent tabular foundation models except TabFM and EXAONE-Tabular.

The overall pairwise win-rate matrix in
Figure~\ref{app:tradeoff-pareto-overall} provides an additional view of
its relative performance across the full benchmark.
Together with its second-place result on regression, these results show
that \ours remains highly competitive across diverse TabArena tasks,
with regression emerging as its strongest regime.

\subsubsection{Performance--Efficiency Trade-offs}

Figure~\ref{fig:tradeoff-pareto}(a) further examines the trade-off between
regression performance and inference efficiency. Improvability
(y-axis, lower is better) measures the relative performance gap to the
best-performing method on each dataset, while the x-axis reports median
inference time per 1K samples.

\ours achieves an improvability of only \textbf{1.7\%} with an inference
time of \textbf{3.12 s} per 1K samples, placing it in the high-performance
region of the trade-off space. Compared with TabFM, which achieves a
slightly lower improvability of 1.6\%, \ours requires \textbf{68\% less
prediction time} (3.12 s vs.\ 9.67 s). Table~\ref{tab:tabarena_reg} further
shows that \ours requires \textbf{82\% less training time}
(6.99 s vs.\ 38.85 s).

Compared with TabPFN-3 at a similar model scale, \ours also achieves
stronger regression performance, improving Elo from 1800 to 1900 and
reducing improvability from 2.4\% to 1.7\%. Overall, these results show
that \ours achieves a favorable performance--efficiency trade-off,
approaching the strongest regression performance on TabArena while
requiring substantially less computation than TabFM.

\begin{figure}[ht!]
    \centering
    \includegraphics[width=\linewidth]{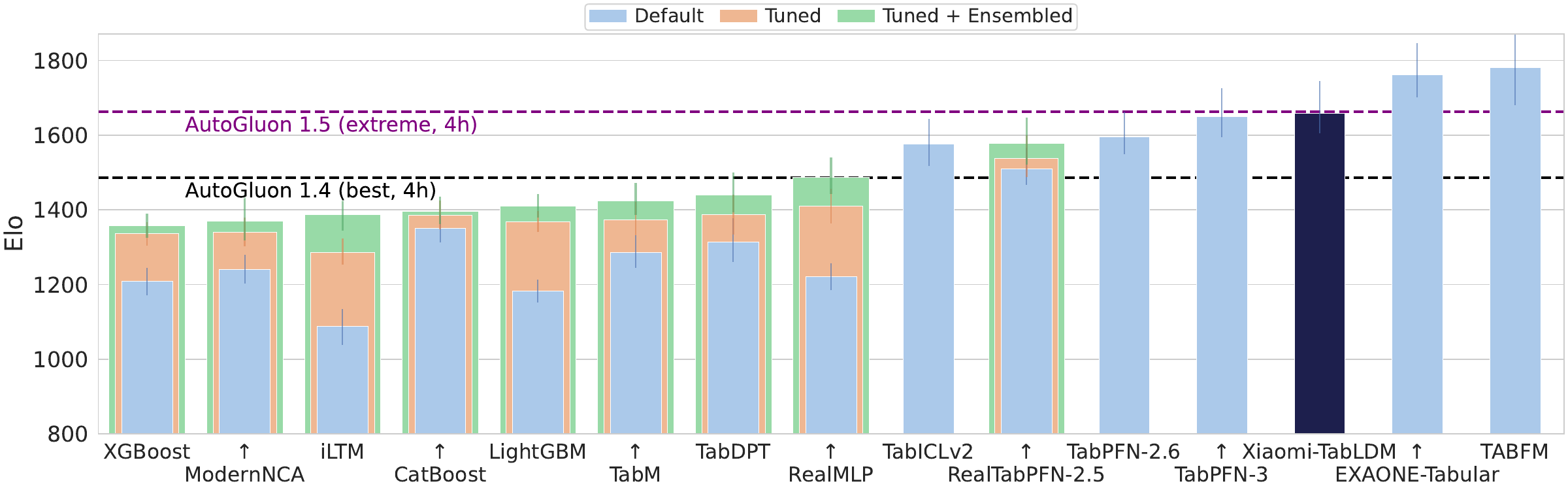}
\caption{
Overall Elo performance on TabArena (higher is better).
}
    \label{fig:tA-overall}
\end{figure}

\begin{table*}[t]
\centering
\caption{
BCCO mean-rank performance across regression, multiclass classification, and
overall evaluations.
Lower values are better.
Numbers in parentheses denote the ranking position under each metric.
}
\label{tab:reg_multicls_overall_results}

\scriptsize
\setlength{\tabcolsep}{1pt}
\renewcommand{\arraystretch}{1.80}

\begin{tabular}{lccc|ccc|ccc}
\toprule
\multirow{2}{*}{\textbf{Model}}
& \multicolumn{3}{c|}{\textbf{Reg}}
& \multicolumn{3}{c|}{\textbf{Multi-Cls}}
& \multicolumn{3}{c}{\textbf{Overall}} \\
\cmidrule(lr){2-4}
\cmidrule(lr){5-7}
\cmidrule(lr){8-10}

& \textbf{Accuracy $\downarrow$}
& \textbf{AUC $\downarrow$}
& \textbf{LogLoss $\downarrow$}
& \textbf{Accuracy $\downarrow$}
& \textbf{AUC $\downarrow$}
& \textbf{LogLoss $\downarrow$}
& \textbf{Accuracy $\downarrow$}
& \textbf{AUC $\downarrow$}
& \textbf{LogLoss $\downarrow$} \\
\midrule

TabFM
& \textbf{2.680 (1)}
& \textbf{2.680 (1)}
& \textbf{2.680 (1)}
& \textbf{2.871 (1)}
& \textbf{2.571 (1)}
& \textbf{2.286 (1)}
& \textbf{3.694 (1)}
& \textbf{3.095 (1)}
& \textbf{3.064 (1)} \\

\textbf{\ours}
& \underline{2.940 (2)}
& \underline{2.940 (2)}
& \underline{2.940 (2)}
& \underline{4.171 (2)}
& 4.257 (4)
& \underline{3.971 (2)}
& 4.451 (3)
& 4.154 (3)
& 3.819 (3) \\

EXAONE-Tabular
& 3.400 (3)
& 3.400 (3)
& 3.400 (3)
& 4.757 (4)
& 4.186 (3)
& 4.086 (4)
& \underline{4.436 (2)}
& \underline{3.731 (2)}
& \underline{3.625 (2)} \\

TabPFN-v3
& 3.560 (4)
& 3.560 (4)
& 3.560 (4)
& 5.129 (5)
& 4.343 (5)
& 4.629 (5)
& 4.999 (4)
& 4.521 (4)
& 4.407 (4) \\

TabICL-v2
& 4.780 (5)
& 4.780 (5)
& 4.780 (5)
& 4.329 (3)
& \underline{4.114 (2)}
& 4.029 (3)
& 5.336 (5)
& 4.805 (5)
& 4.744 (5) \\

LimiX
& 6.500 (6)
& 6.500 (6)
& 6.500 (6)
& 5.143 (6)
& 5.514 (6)
& 5.057 (6)
& 6.321 (6)
& 6.071 (6)
& 5.915 (6) \\

\midrule

RealMLP
& 8.330 (7)
& 8.330 (7)
& 8.330 (7)
& 9.614 (10)
& 10.029 (12)
& 9.714 (11)
& 7.229 (7)
& 8.623 (8)
& 8.481 (8) \\

CatBoost
& 8.960 (8)
& 8.960 (8)
& 8.960 (8)
& 8.800 (8)
& 9.000 (9)
& 8.229 (7)
& 9.400 (11)
& 8.591 (7)
& 8.249 (7) \\

FT-Transformer
& 9.030 (9)
& 9.030 (9)
& 9.030 (9)
& 9.257 (9)
& 8.943 (8)
& 8.886 (10)
& 8.121 (8)
& 8.815 (9)
& 8.555 (9) \\

MLP-PLR
& 9.640 (10)
& 9.640 (10)
& 9.640 (10)
& 9.700 (11)
& 9.986 (11)
& 10.171 (12)
& 8.346 (10)
& 9.414 (11)
& 9.509 (11) \\

ModernNCA
& 9.840 (11)
& 9.840 (11)
& 9.840 (11)
& 8.657 (7)
& 8.343 (7)
& 8.657 (8)
& 8.220 (9)
& 9.109 (10)
& 9.245 (10) \\

LightGBM
& 10.920 (12)
& 10.920 (12)
& 10.920 (12)
& 10.400 (13)
& 10.971 (13)
& 12.543 (13)
& 10.602 (12)
& 10.646 (13)
& 11.918 (13) \\

XGBoost
& 11.120 (13)
& 11.120 (13)
& 11.120 (13)
& 10.229 (12)
& 9.657 (10)
& 8.800 (9)
& 10.933 (13)
& 10.400 (12)
& 9.961 (12) \\

KNN
& 13.300 (14)
& 13.300 (14)
& 13.300 (14)
& 11.943 (14)
& 13.086 (14)
& 13.943 (14)
& 12.914 (14)
& 13.027 (14)
& 13.508 (14) \\

\bottomrule
\end{tabular}
\vspace{2em}
\end{table*}
\subsection{BCCO}\label{BCCO}

We further evaluate \ours on the Balanced Comprehensive Challenging Omni-domain
(BCCO) benchmark~\cite{zhang2025limix}, which consists of two subsets,
BCCO-CLS for classification and BCCO-REG for regression. BCCO is constructed
from a broad collection of open-source structured datasets after extensive
deduplication and cleaning, and is designed to cover heterogeneous real-world
prediction settings. Compared with commonly used tabular benchmarks, BCCO
contains greater variation in dataset characteristics, including sample size,
feature dimensionality, number of classes, categorical-to-numerical feature
ratio, sample-to-feature ratio, and missing-value rate. Notably, nearly
one-third of the datasets contain missing values, making the benchmark
particularly challenging for models that need to generalize across diverse
data conditions.

In our evaluation, BCCO-CLS contains 106 classification datasets and BCCO-REG
contains 50 regression datasets. The benchmark excludes extremely large
datasets with more than 50,000 training samples, 10,000 features, or 10 target
classes. Its broad coverage across dataset characteristics enables evaluation
over a diverse range of task regimes rather than concentrating on a small
subset of dataset types.
\begin{figure}[t!]
    \centering
    \includegraphics[width=\linewidth]{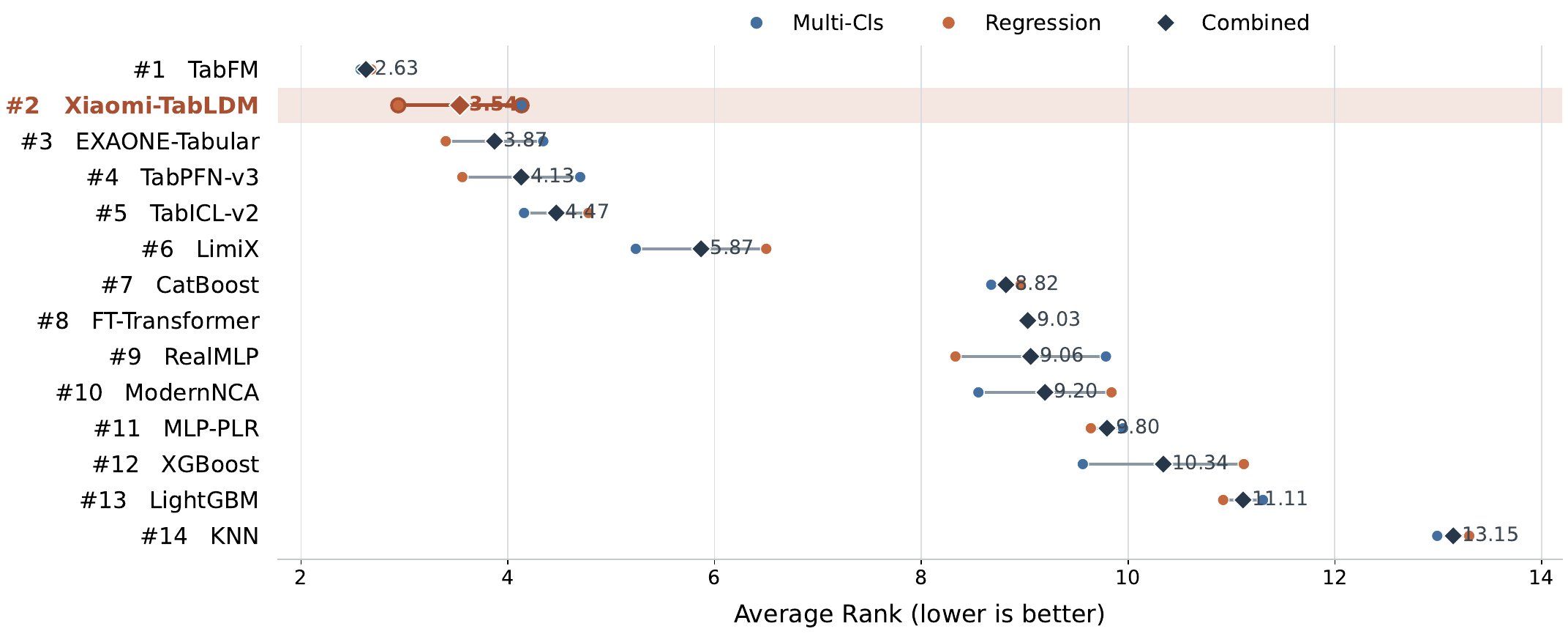}
\caption{
Average-rank comparison on the BCCO benchmark.
Circles denote the average ranks on BCCO-CLS and BCCO-REG, while
diamonds denote the combined average rank across the two settings.
Models are ordered by the combined average rank; lower is better.
}
    \label{fig:bcco_acc}
\end{figure}
\paragraph{Main Results.}
Table~\ref{tab:reg_multicls_overall_results} reports the mean-rank performance
on BCCO across regression, multiclass classification, and the overall
evaluation. \ours demonstrates particularly strong performance on regression,
ranking second under all three metrics, with a mean rank of 2.940 for
Accuracy, AUC, and LogLoss. It consistently outperforms other strong tabular
foundation models, including EXAONE-Tabular, TabPFN-v3, TabICL-v2, and LimiX,
and is surpassed only by TabFM.

On multiclass classification, \ours remains highly competitive, ranking second
under both Accuracy (4.171) and LogLoss (3.971), while achieving fourth place
under AUC (4.257). When aggregating across the full BCCO benchmark, \ours
consistently ranks third under Accuracy (4.451), AUC (4.154), and LogLoss
(3.819). Overall, these results show that \ours achieves robust top-tier
performance across heterogeneous BCCO tasks, with a particularly clear
advantage on regression while maintaining competitive performance on
multiclass classification.

\subsection{OpenML-CTR23}\label{ctr23}

We further evaluate \ours on the OpenML Curated Tabular Regression
benchmarking suite 2023 (OpenML-CTR23), a benchmark specifically designed
for tabular regression. The original suite contains 35 regression problems
selected according to a set of strict curation criteria, providing a
standardized and reliable testbed for comparing regression methods across
diverse tabular datasets. In our experiments, we report results on 33
datasets and compare \ours against both tabular foundation models and
conventional learning baselines.

\paragraph{Main Results.}
Figure~\ref{fig:ctr23} reports the average-rank results on OpenML-CTR23~\cite{fischer2023openml}.
\ours achieves the best average rank of \textbf{3.03}, ranking first
among all evaluated methods.
It also consistently ranks ahead of other strong tabular foundation models,
including TabPFN-v3 (3.71), TabICL-v2 (4.55), and LimiX (5.91).
These results demonstrate the strong regression capability of \ours and its
ability to generalize consistently across the diverse tasks covered by
OpenML-CTR23.

\subsection{Embeddings.}
\begin{figure}[htbp]
    \centering
    \includegraphics[width=0.9\linewidth]{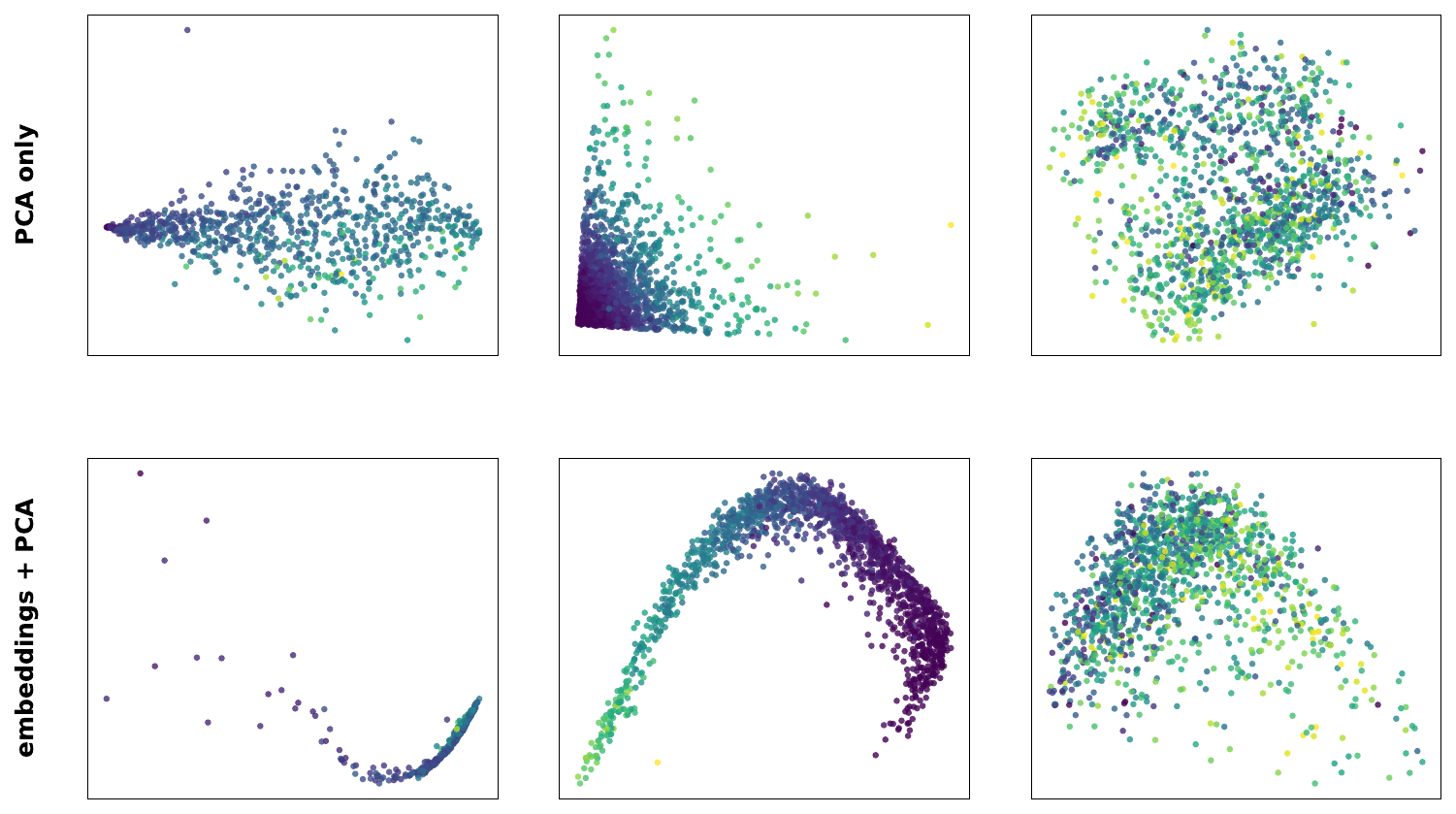}
    \caption{PCA visualization of the representations learned by \ours on three classification datasets. The top row presents projections of the original feature space, while the bottom row presents projections of the corresponding row embeddings extracted from \ours. Points denote individual samples and colors denote class labels. Compared with the raw inputs, the learned embeddings exhibit more organized structures that better reflect the class distribution.}
    \label{fig:embedding}
\end{figure}

Finally, we examine whether \ours learns meaningful representations of tabular samples. We extract the row embeddings produced by \ours and apply PCA to project them into two dimensions for visualization. Figure~\ref{fig:embedding} compares PCA applied directly to the original input features with PCA applied to the learned embeddings across three representative datasets. While the original feature space exhibits relatively diffuse and weakly organized structures, the embeddings learned by \ours reveal substantially clearer low-dimensional organization, including coherent curved manifolds and more compact sample structures. This suggests that \ours transforms heterogeneous tabular inputs into representations that better capture the underlying structure of the data.

As shown in Figure~\ref{fig:embedding}, \ours learns structured row embeddings from tabular data. The upper plots show 2D PCA applied directly to the original features of three datasets, while the lower plots show PCA applied to the corresponding row embeddings produced by \ours. The learned representations exhibit substantially clearer and more coherent low-dimensional structures than the original feature space.

\section{Conclusion}
In this report, we present \ours, a tabular foundation model designed to achieve strong predictive performance while maintaining an efficient model and inference profile. Across four public benchmark suites covering diverse classification and regression tasks, \ours consistently ranks among the strongest evaluated methods. Its performance is particularly strong on regression, where it achieves top-tier results across TALENT, TabArena, BCCO, and OpenML-CTR23, while remaining competitive across classification settings.

Beyond predictive accuracy, our evaluation also highlights a favorable performance--efficiency trade-off. At a model scale comparable to TabPFN-3, \ours achieves stronger performance in several evaluation regimes, while requiring substantially less computational cost than larger or heavily tuned alternatives such as TabFM and AutoGluon in relevant comparisons. These results indicate that improving tabular foundation models does not necessarily require scaling model size or computational cost alone; effective architectural design and capacity allocation can provide another practical path toward stronger and more efficient tabular learning systems.

The results demonstrate that \ours provides a competitive balance of predictive performance, model capacity, and computational efficiency across heterogeneous tabular tasks. We hope these findings provide a useful basis for further exploration of scalable and efficient foundation-model architectures for structured data.

\newpage
\small
\bibliographystyle{plain}
\bibliography{refs}

\newpage
\appendix
\newpage
\section*{Appendix}

\section{Contributions and Acknowledgments}

We express our sincere gratitude to all contributors for their dedication to the design,
implementation, experimentation, and documentation of Xiaomi-TabLDM.
Their collaborative efforts across data construction, model architecture, training,
system integration, and empirical evaluation were instrumental to the success of this work
and the release of the accompanying model and codebase.
The contributors to this work are listed as follows:




\noindent
\textbf{Core Contributors:} Penghui Wang$^{\dagger}$, Wei Liu$^{\dagger}$, Hong Wang, Chengyue Huang, Yuxi Sun, Zirui Wang, Hongming Huang, Quan Wang, Chunxiao Liu$^{*}$, Erli Meng, Bin Wang

\textbf{Contributors:} Zhenwei Xin, Ping Hou, Jie Yu

\vspace{0.5em}
\noindent
$^{\dagger}$ Equal contribution\\
$^{*}$ Corresponding author






\subsection{Architectural Hyperparameters}
\label{app:arch-hyperparams}
The released \ours classifier and regressor checkpoints share all architectural hyperparameters and the sparse MoE configuration, differing only in the task-specific target encoder and the
  final linear layer of the output decoder. These hyperparameters are summarized in the tables below (Tables~\ref{tab:moe-feat-embed}--\ref{tab:moe-ffn}).
  \begin{table}[htbp]
  \centering
  \caption{Column-wise feature embedding.}
  \label{tab:moe-feat-embed}
  \begin{tabular}{@{}llp{7.2cm}@{}}
    \toprule
    \textbf{Hyperparameter} & \textbf{Value} & \textbf{Description} \\
    \midrule
    \texttt{embed\_dim}              & 128 & Base embedding dimension used model-wide \\
    \texttt{col\_feature\_group\_size} & 3 & Features per circular-shift group \\
    \texttt{global\_dilation}      & \texttt{adaptive} &  Offset scheme for the second column stream  \\
    \texttt{global\_max\_span}     & 32  & Target span $s_{\max}$ for the second stream \\
    \texttt{col\_num\_blocks}        & 3   & Induced self-attention blocks \\
    \texttt{col\_nhead}              & 8   & Attention heads per block \\
    \texttt{col\_num\_inds}          & 128 & Inducing points per column \\
    \bottomrule
  \end{tabular}
\end{table}

\begin{table}[htbp]
  \centering
  \caption{Row-wise feature aggregation.}
  \label{tab:moe-feat-agg}
  \begin{tabular}{@{}llp{7.2cm}@{}}
    \toprule
    \textbf{Hyperparameter} & \textbf{Value} & \textbf{Description} \\
    \midrule
    \texttt{row\_num\_blocks}  & 3        & Transformer blocks \\
    \texttt{row\_nhead}        & 8        & Attention heads per block \\
    \texttt{row\_num\_cls}     & 4        & CLS tokens aggregated per row \\
    \texttt{use\_rope}         & True     & Rotary positional embeddings (RoPE) enabled \\
    \texttt{row\_rope\_base}   & $100\,000$ & RoPE base frequency $\theta$ \\
    \texttt{block\_size}           & 4   & AttnRes residual block size \\
    \texttt{attnres\_stride}       & 4   & Depth stride at which AttnRes is applied \\
    \bottomrule
  \end{tabular}
\end{table}

\begin{table}[htbp]
  \centering
  \caption{Dataset-wise in-context learning.}
  \label{tab:moe-icl}
  \begin{tabular}{@{}llp{7.2cm}@{}}
    \toprule
    \textbf{Hyperparameter} & \textbf{Value} & \textbf{Description} \\
    \midrule
    \texttt{icl\_emsize} (derived) & 512 & \texttt{embed\_dim} $\times$ \texttt{row\_num\_cls} $= 128 \times 4$ \\
    \texttt{icl\_num\_blocks}      & 24  & Transformer blocks \\
    \texttt{icl\_nhead}            & 8   & Attention heads per block \\
    \texttt{ff\_factor}            & 2   & Feed-forward expansion factor \\
    \texttt{block\_size}           & 4   & AttnRes residual block size \\
    \texttt{attnres\_stride}       & 4   & Depth stride at which AttnRes is applied \\
    \bottomrule
  \end{tabular}
\end{table}

  \begin{table}[htbp]
    \centering
    \caption{The output decoder uses a two-layer MLP with a
GELU activation~\cite{hendrycks2016gelu}.
    \texttt{Linear}$(512,1024)\to\mathrm{GELU}\to\texttt{Linear}(1024,\texttt{out\_dim})$.
    Only \texttt{out\_dim} and the target encoder differ between the two models.}
    \label{tab:moe-decoder}
    \begin{tabular}{@{}l >{\centering\arraybackslash}p{5.0cm} >{\centering\arraybackslash}p{5.0cm}@{}}
      \toprule
      \textbf{Hyperparameter} & \textbf{Classifier} & \textbf{Regressor} \\
      \midrule
      \texttt{out\_dim}       & 10 & 999 \\
      target encoder          & \texttt{OneHotAndLinear}$(10,512)$ & \texttt{Linear}$(1,512)$ \\
      \bottomrule
    \end{tabular}
  \end{table}

\begin{table}[htbp]
    \centering
    \caption{Sparse MoE. Replaces the dense FFN in selected ICL blocks.
    Each expert is a 2-layer MLP with the same hidden width as the dense FFN it
    replaces, \texttt{icl\_emsize} $\times$ \texttt{ff\_factor} $= 512 \times 2 = 1024$.}
    \label{tab:moe-ffn}
    \begin{tabular}{@{}llp{6.6cm}@{}}
      \toprule
      \textbf{Hyperparameter} & \textbf{Value} & \textbf{Description} \\
      \midrule
      \texttt{icl\_moe\_layers}              & \texttt{last\_8} & MoE blocks; final 8 of 24 \\
      \texttt{icl\_moe\_num\_experts}        & 2    & Routed experts per layer \\
      \texttt{icl\_moe\_top\_k}              & 1    & Experts activated per token \\
      \texttt{icl\_moe\_num\_shared\_experts}& 1    & Always-on shared expert \\
      \texttt{expert\_hidden\_dim} (derived) & 1024 & Per-expert hidden width, \texttt{icl\_emsize} $\times$ \texttt{ff\_factor} \\
      \texttt{icl\_moe\_init\_from\_dense}   & True & Experts initialized from the frozen dense FFN  \\
      \texttt{icl\_moe\_router\_jitter}      & 0.0  & Router input jitter (disabled) \\
      \bottomrule
    \end{tabular}
  \end{table}
\subsection{MoE auxiliary losses}
\label{app:moe aux_loss}
We regularize routing with two auxiliary terms per MoE layer: a Switch-style load-balance loss\cite{fedus2022switch} that spreads tokens evenly over the $M_R$ routed experts, and a router $z$-loss\cite{zoph2022st} that keeps the logits bounded. Let $z_{t,i}$ be the router logit of expert $i$ on row token $t$ and $T$ is the total number of row tokens,
  \begin{equation}
    \mathcal{L}_{\mathrm{bal}}=M_R\sum_{i=1}^{M_R} f_i\,P_i,
    \qquad
    \mathcal{L}_{z}=\frac{1}{T}\sum_{t=1}^{T}\Bigl(\log\!\sum_{i=1}^{M_R}e^{z_{t,i}}\Bigr)^{2}
 \end{equation}
  where $f_i$ is the fraction of tokens routed to expert $i$ and $P_i$ its mean
  gate probability, so $\mathcal{L}_{\mathrm{bal}}$ is minimized under uniform
  load. Summed over MoE layers and scaled by the coefficients in
  Table~\ref{tab:moe-aux}, they are added to the task loss
  $\mathcal{L}_{\mathrm{task}}$ (cross-entropy for the classifier, pinball loss for the regressor),
  \begin{equation}
    \mathcal{L}=\mathcal{L}_{\mathrm{task}}
    +\lambda(\alpha_{\mathrm{bal}}\,\mathcal{L}_{\mathrm{bal}}
    +\alpha_{z}\,\mathcal{L}_{z})
  \end{equation}
\begin{table}[htbp]
    \centering
    \caption{MoE auxiliary losses. Both terms are computed per MoE layer,
    summed over layers, scaled by \texttt{icl\_moe\_aux\_loss\_weight}, and added to
    the task loss.}
    \label{tab:moe-aux}
    \begin{tabular}{@{}lllp{5.2cm}@{}}
      \toprule
      \textbf{Hyperparameter} & \textbf{Symbol} & \textbf{Value} & \textbf{Description} \\
      \midrule
      \texttt{icl\_moe\_router\_z\_loss\_coef}     & $\alpha_{z}$            & \texttt{1e-3} & Router z-loss weight \\
      \texttt{icl\_moe\_load\_balance\_loss\_coef} & $\alpha_{\mathrm{bal}}$ & \texttt{1e-2} & Load-balance weight \\
      \texttt{icl\_moe\_aux\_loss\_weight}         & $\lambda$               & \texttt{1.0}  & Auxiliary loss multiplier \\
      \bottomrule
    \end{tabular}
  \end{table}
\subsection{Model parameter counts}
As shown in Table~\ref{tab:moe-params}, we list the parameter counts of the \ours classification and regression models. Total parameters count all weights, while active parameters count those actually used in a single forward pass, determined by the sparse MoE configuration.
  \begin{table}[htbp]
    \centering
    \caption{Parameter counts of the released \ours checkpoints.}
    \label{tab:moe-params}
    \begin{tabular}{@{}llcc@{}}
      \toprule
      \textbf{Model} & \textbf{Type} & \textbf{Total params} & \textbf{Active params} \\
      \midrule
      classifier & Classification (\texttt{max\_classes}=10) & 70.08\,M & 61.67\,M \\
      regressor  & Regression (\texttt{quantiles}=999)                     & 71.08\,M & 62.68\,M \\
      \bottomrule
    \end{tabular}
  \end{table}

\subsection{Many-class classification}
  \label{app:many-class}

 \ours is pretrained with at most $\text{max\_classes}=10$ classes. For tasks with $C>10$ classes, the mixed-radix ensembling of
TabICLv2\cite{qu2026tabiclv2betterfasterscalable} is applied: the labels are re-encoded into $D$ mixed-radix digits over balanced bases, and the column-wise transformer is run once per digit view with the outputs averaged, thereby supporting an arbitrary number of classes without retraining.
Formally, balanced bases $[k_0,\dots,k_{D-1}]$ are chosen with every $k_i\le 10$ and $\prod_{i=0}^{D-1} k_i \ge C$, and each label is rewritten into $D$ mixed-radix
  digits,
  \begin{equation}
    y^{(i)} = \left\lfloor y \Big/ \textstyle\prod_{j>i} k_j \right\rfloor \bmod k_i,
    \qquad i = 0,\dots,D-1.
  \end{equation}
 Mixed-radix ensembling acts at the column-wise feature embedding stage, while hierarchical classification at the dataset-wise ICL stage handles the remaining many-class structure.

\clearpage
\begin{figure}[t!]
    \begin{subfigure}[b]{1\linewidth}
        \centering
        \includegraphics[width=1\linewidth]{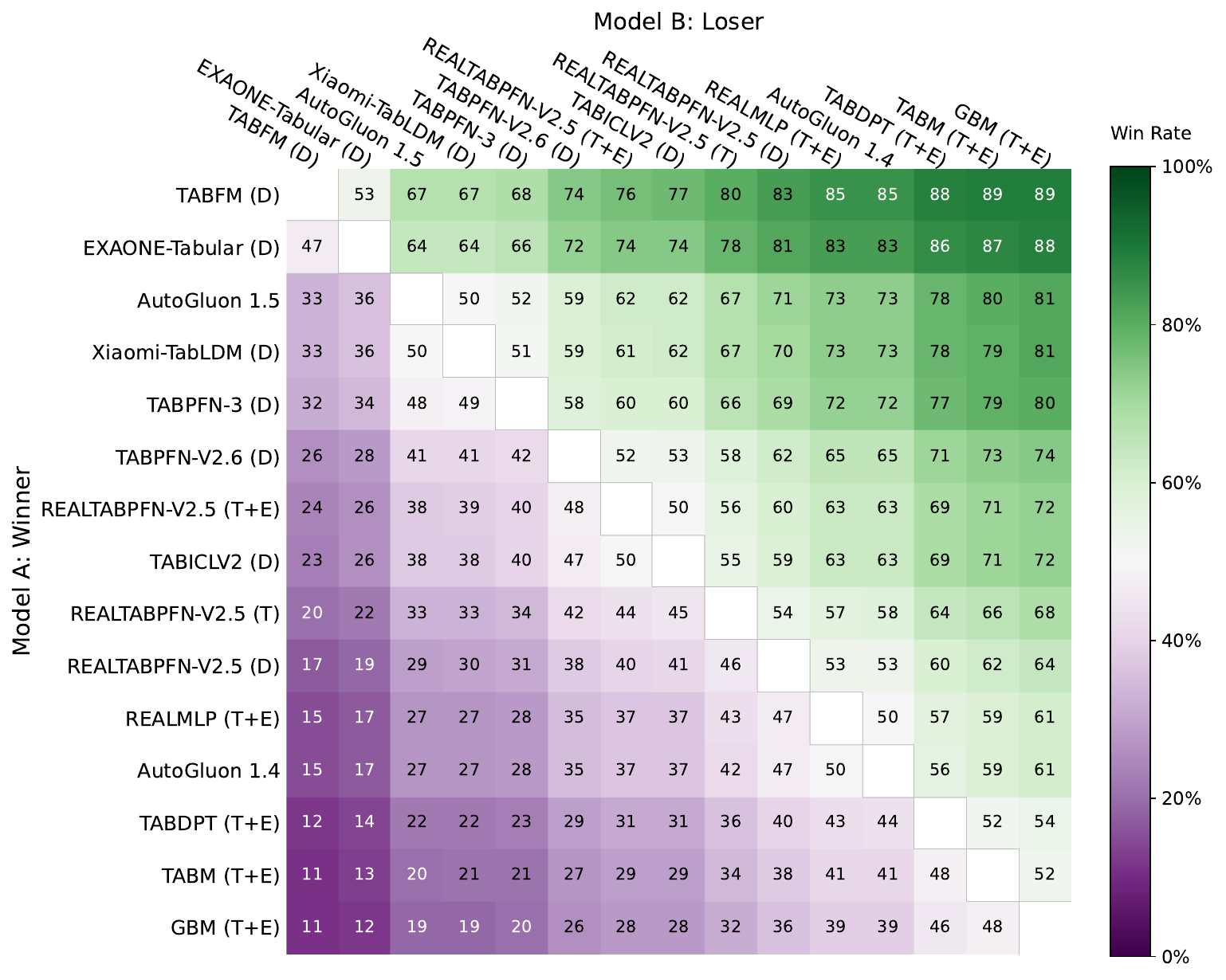}
        \label{fig:winrate-regression}
    \end{subfigure} 
\caption{
Pairwise win-rate matrix on the full TabArena benchmark.
Each entry $(i,j)$ reports the proportion of instances on which model $i$
outperforms model $j$. Diagonal entries denote self-comparison
(50\% by definition).
}
    \label{app:tradeoff-pareto-overall}
\end{figure}

\section{Results}

Our comparisons cover a broad range of tabular learning approaches.
For tabular foundation models, we include
TabFM~\cite{kong2026tabfm},
EXAONE-Tabular~\cite{eo2026exaonetabular},
TabPFN-3~\cite{grinsztajn2026tabpfn3technicalreport},
TabPFN-2.6~\cite{priorlabs2026tabpfn26},
RealTabPFN~\cite{garg2025realtabpfn},
TabICLv2~\cite{qu2026tabiclv2betterfasterscalable},
TabDPT~\cite{ma2025tabdpt},
LimiX~\cite{zhang2025limix},
iLTM~\cite{bonet2026iltm},
TabSTAR~\cite{arazi2025tabstar},
TabFlex~\cite{zeng2025tabflex},
and OrionMSP~\cite{bouadi2025orionmsp}.
We further compare against strong deep and feature-learning methods,
including RealMLP~\cite{holzmuller2024betterbydefault},
ModernNCA~\cite{ye2025modernnca},
TabM~\cite{gorishniy2025tabm},
and xRFM~\cite{beaglehole2026xrfm},
as well as tree-based and classical baselines such as
XGBoost~\cite{chen2016xgboost},
CatBoost~\cite{dorogush2018catboost},
LightGBM~\cite{ke2017lightgbm},
and EBM~\cite{nori2019interpretml}.
AutoGluon~\cite{erickson2020autogluon} is included as a strong AutoML
baseline. The complete set of evaluated methods and configurations,
including additional classical and neural baselines, follows the
TabArena benchmark~\cite{erickson2025tabarena} and is reported in the
appendix.

\subsection{Details of TabArena}

\subsubsection{Evaluation Metrics}

We reuse the official TabArena~\cite{erickson2025tabarena} evaluation metrics and evaluation code for generating the TabArena plots and leaderboard tables.

\textbf{Elo.}
Following TabArena, we evaluate models using the Elo rating system. Elo is based on pairwise model comparisons, where the difference between two models' ratings determines their expected win probability. A 400-point Elo difference corresponds to an expected win probability of approximately 91\% for the higher-rated model. Following the official TabArena protocol, we calibrate an Elo score of 1000 to the default RandomForest configuration and perform 200 bootstrap rounds to estimate 95\% confidence intervals. For task-level comparisons, TabArena uses ROC-AUC for binary classification, log-loss for multiclass classification, and RMSE for regression.

\textbf{Improvability.}
We additionally report the Improvability metric introduced in TabArena. For a dataset $i$, it measures the relative reduction in error required for a method to match the best-performing method on that dataset:
\begin{equation}
    \mathrm{Improvability}_i
    =
    \frac{\mathrm{err}_i - \mathrm{best\_err}_i}
         {\mathrm{err}_i}
    \times 100\%.
\end{equation}
The final Improvability score is averaged across datasets. Lower values are better, with $0\%$ indicating performance equal to the best observed method on every dataset.

\textbf{Dataset wins.}
We also report the number of dataset wins, which measures how often a method achieves the best performance among the compared methods. Together with Elo and Improvability, this provides a complementary view of both average performance and task-level dominance.

\subsubsection{Tuning and Efficiency Plots}

We follow the official TabArena plotting protocol to analyze the effect of tuning and ensembling. For methods with multiple configurations, the plots compare the default configuration with tuned and tuned-and-ensembled variants. The tuning trajectories are constructed from increasingly large ensembles of sampled configurations, following the TabArena evaluation procedure.

In addition to predictive performance, we report computational efficiency using the median training and inference time per 1K samples. These results allow us to compare not only the achievable performance of different methods, but also the additional computational cost introduced by dataset-specific tuning and ensembling. In particular, they highlight the trade-off between strong default performance and substantially more expensive tuned pipelines.




\subsubsection{TabArena Leaderboard Tables}

Tables~\ref{tab:elo_overall_all} and~\ref{tab:elo_all} report the detailed
TabArena leaderboard results on the full benchmark and the regression subset,
respectively.

Across all 51 datasets and 816 tasks, \ours achieves an Elo score of
\textbf{1659}, ranking \textbf{fourth} by Elo point estimate.
Its performance is nearly tied with AutoGluon 1.5 extreme (1662) and
surpasses strong recent tabular foundation models including TabPFN-3 (1650),
TabPFN-2.6 (1596), RealTabPFN-2.5 with tuning and ensembling (1579), and
TabICLv2 (1576). \ours also achieves \textbf{3.1 dataset wins}, exceeding
TabPFN-3 and the other recent tabular foundation models except TabFM and
EXAONE-Tabular. These results demonstrate that \ours remains highly
competitive across the full TabArena benchmark.

The strongest relative performance of \ours is observed on regression.
Across the 13 regression datasets, \ours reaches an Elo score of
\textbf{1900}, ranking \textbf{second} behind only TabFM (2019).
It outperforms EXAONE-Tabular (1885), TabPFN-3 (1800),
AutoGluon 1.5 extreme (1776), TabPFN-2.6 (1741),
RealTabPFN-2.5 with tuning and ensembling (1736), and TabICLv2 (1679).
In addition, \ours achieves \textbf{1.8 dataset wins} and an improvability
of only \textbf{1.7\%}, the second-best result behind TabFM (1.6\%).

Notably, this regression performance is achieved with substantially lower
computational cost than TabFM. \ours requires 6.99~s of training time and
3.12~s of prediction time per 1K samples, compared with 38.85~s and
9.67~s for TabFM, corresponding to approximately \textbf{82\% less training
time} and \textbf{68\% less prediction time}. Overall, the TabArena
leaderboards show that \ours combines competitive performance across the
full benchmark with particularly strong regression performance and a
favorable performance--efficiency trade-off.

\subsection{Details on TALENT benchmark}
We evaluate \ours on \tL (Tabular Analytics and LEarNing Toolbox)~\cite{liu2025talent}, a comprehensive benchmark and evaluation framework for tabular prediction. TALENT brings together a broad range of classical machine-learning and deep-learning methods under a unified evaluation pipeline, with standardized preprocessing and model interfaces to facilitate consistent comparison across heterogeneous tabular datasets. The benchmark covers both classification and regression tasks with substantial variation in dataset size, feature dimensionality, class structure, and feature composition.

For classification, \tL supports multiple evaluation metrics, including Accuracy, F1-score, Log Loss, and AUC. In our experiments shown in Figure~\ref{fig:talent_results}, we use \textbf{Accuracy} as the primary classification metric and compute the average rank of each model across datasets based on its per-dataset Accuracy. Thus, the reported classification rank reflects how consistently a model performs relative to other methods across the benchmark, rather than its average Accuracy value alone.

\subsection{Prior visualizations}

\begin{figure}[htbp]
    \centering
    \includegraphics[width=0.8\linewidth]{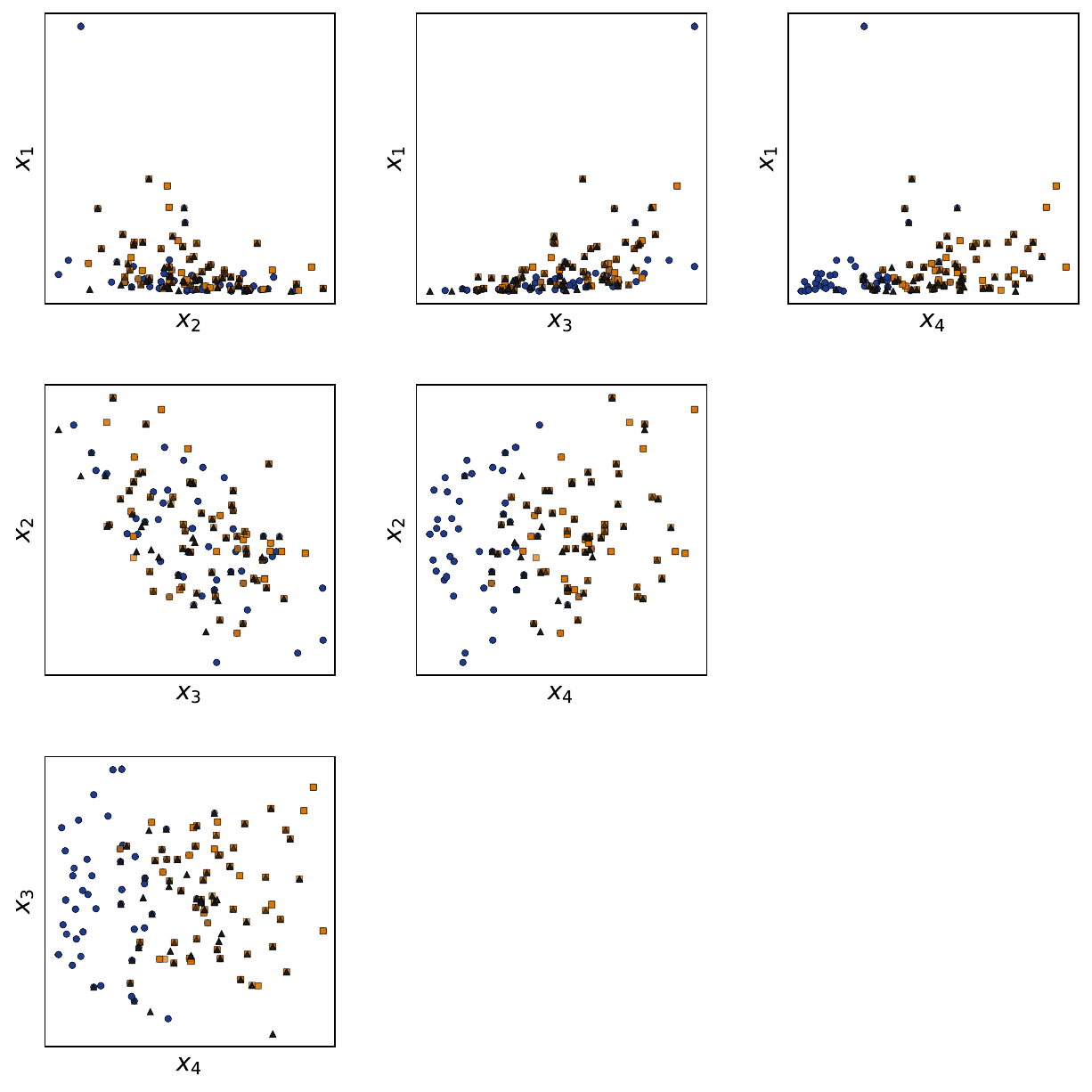}
    \caption{Example classification dataset sampled from the prior with four covariates. Each subplot in row $i$, column $j$ visualizes covariates $i$ and $j+1$, with colors denoting the target classes.}
    \label{fig:placeholder}
\end{figure}

\begin{figure}[htbp]
    \centering

    \begin{subfigure}{\linewidth}
        \centering
        \includegraphics[width=0.9\linewidth]{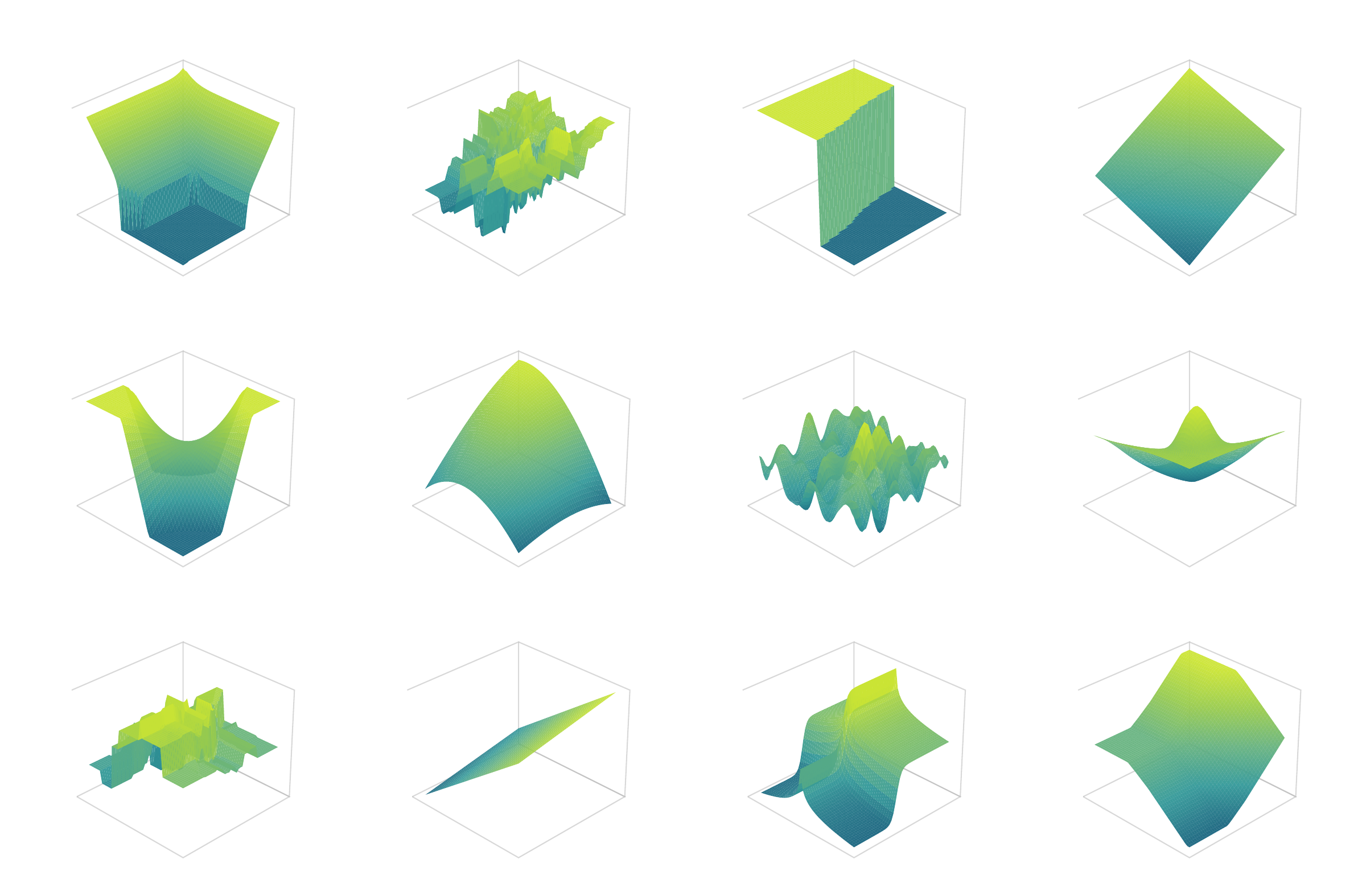}
        \caption{Examples of individual edge functions generated by the new combiner mechanisms.}
        \label{fig:individual_edge_functions}
    \end{subfigure}

    \vspace{0.5em}

    \begin{subfigure}{\linewidth}
        \centering
        \includegraphics[width=0.9\linewidth]{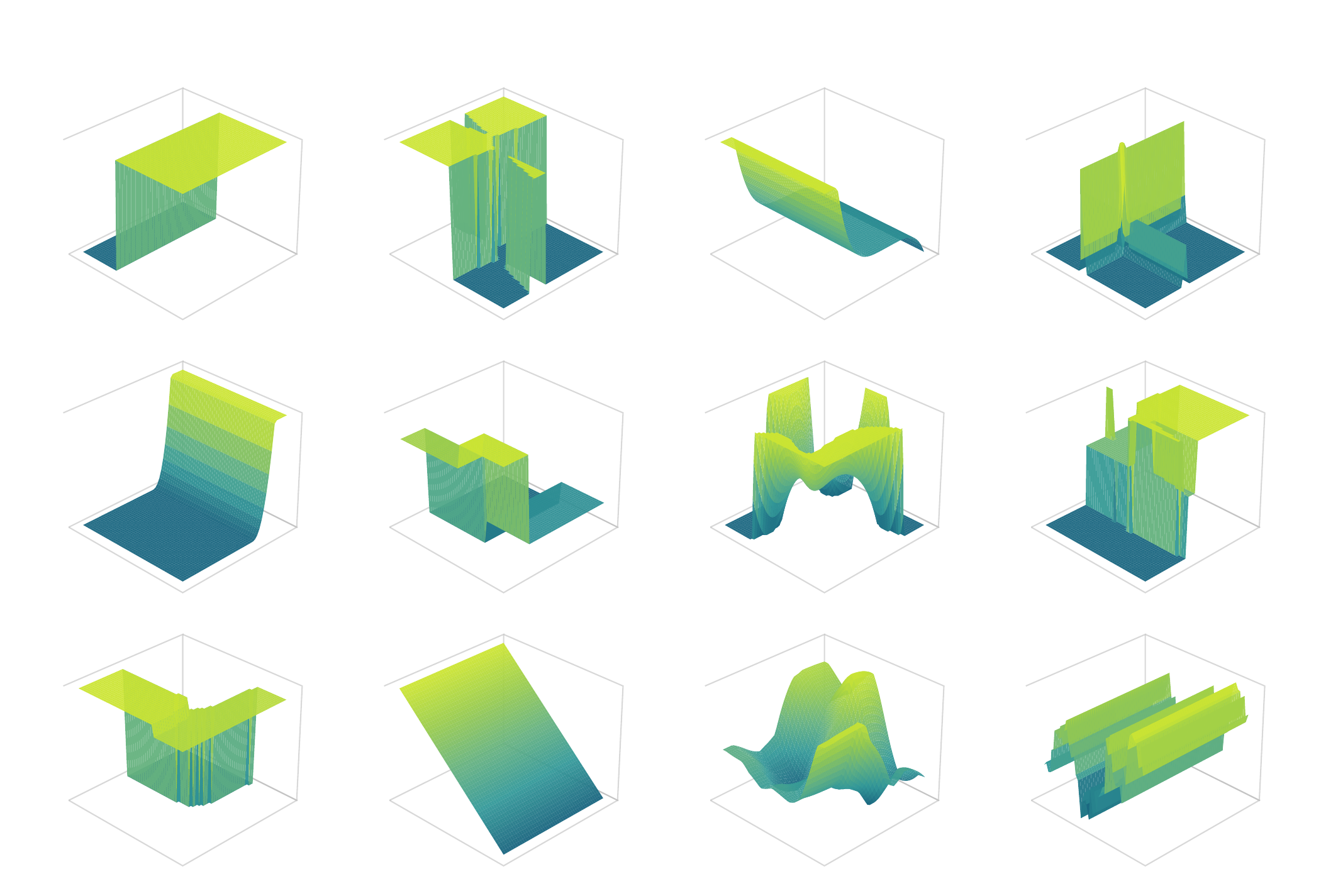}
        \caption{Examples of more complex functional relationships obtained by composing and mixing multiple edge functions.}
        \label{fig:composed_edge_functions}
    \end{subfigure}

    \caption{
    Visualization of functional relationships generated by the new combiner mechanisms in our SCM prior.
    The top panel shows individual edge functions, while the bottom panel illustrates more complex relationships constructed by composing and mixing these functions.
    Although mechanisms in the prior may have varying input dimensionality, we visualize them on a two-dimensional grid for clarity.
    }
    \label{fig:scm_function_visualization}
\end{figure}

To illustrate the functional diversity introduced by our new combiner mechanisms, 
we visualize representative relationships sampled from the SCM prior in 
Fig.~\ref{fig:scm_function_visualization}. 
The individual edge functions in Fig.~\ref{fig:individual_edge_functions} span 
a broad range of behaviors, including smooth nonlinear mappings, localized 
responses, piecewise transitions, and highly irregular surfaces. 
These functions serve as basic building blocks for constructing dependencies 
between variables in the sampled SCMs.

More importantly, composing and mixing multiple edge functions further expands 
the space of functional relationships represented by the prior. 
As shown in Fig.~\ref{fig:composed_edge_functions}, the resulting functions can 
exhibit substantially richer structures, combining sharp transitions, local 
nonlinearities, and heterogeneous behaviors within a single relationship. 
This compositional design allows the synthetic prior to cover a wider variety of 
dependencies than relying on a small set of predefined functional forms alone. 
Although the actual mechanisms operate on inputs of varying dimensionality, we 
restrict the visualization to two-dimensional input spaces for clarity.

\small
\setlength{\tabcolsep}{4.5pt}
\renewcommand{\arraystretch}{1.02}

\begin{longtable}{llccrr}

\caption{
Overall performance on TabArena.
We compare models in terms of Elo, number of wins, improvability, and training and prediction time. Here \textbf{D} denotes the default (untuned) model, \textbf{T} the fine-tuned model, and \textbf{T+E} ensembling after fine-tuning.
}
\label{tab:elo_overall_all} \\

\toprule
\textbf{Model} & \textbf{Elo ($\uparrow$)} & \textbf{\#wins ($\uparrow$)} & \textbf{Improva-} & \textbf{Train time} & \textbf{Predict time} \\
 &  &  & \textbf{bility ($\downarrow$)} & \textbf{per 1K [s]} & \textbf{per 1K [s]} \\
\midrule
\endfirsthead

\multicolumn{6}{c}{
\textit{Table \thetable{} continued from previous page}
} \\
\toprule
\textbf{Model} & \textbf{Elo ($\uparrow$)} & \textbf{\#wins ($\uparrow$)} & \textbf{Improva-} & \textbf{Train time} & \textbf{Predict time} \\
 &  &  & \textbf{bility ($\downarrow$)} & \textbf{per 1K [s]} & \textbf{per 1K [s]} \\
\midrule
\endhead

\midrule
\multicolumn{6}{r}{
\textit{Continued on next page}
} \\
\endfoot

\bottomrule
\endlastfoot

TABFM (D)
& \textcolor{gold}{\textbf{1782${}_{-102,+109}$}}
& \textcolor{gold}{\textbf{21.1}}
& \textcolor{gold}{\textbf{4.4\%}}
& 38.85
& 10.52 \\

EXAONE-Tabular (D)
& \textcolor{silver}{\textbf{1762${}_{-61,+85}$}}
& \textcolor{silver}{\textbf{5.4}}
& \textcolor{silver}{\textbf{7.9\%}}
& 9.00
& 2.10 \\

AutoGluon 1.5 (extreme, 4h)
& \textcolor{bronze}{\textbf{1662${}_{-62,+75}$}}
& \textcolor{bronze}{\textbf{3.6}}
& \textcolor{bronze}{\textbf{8.3\%}}
& 289.07
& 4.03 \\

\textbf{\ours (D)}
& 1659${}_{-55,+86}$
& 3.1
& 9.7\%
& 10.92
& 4.18 \\

TabPFN-3 (D)
& 1650${}_{-55,+76}$
& 1.8
& 9.6\%
& 4.97
& 0.58 \\

TabPFN-2.6 (D)
& 1596${}_{-47,+68}$
& 0.3
& 11.0\%
& 5.48
& 0.55 \\

RealTabPFN-2.5 (T+E)
& 1579${}_{-58,+70}$
& 0.6
& 10.7\%
& 2040.22
& 8.91 \\

TabICLv2 (D)
& 1576${}_{-58,+67}$
& 1.3
& 10.4\%
& 4.02
& 0.38 \\

RealTabPFN-2.5 (T)
& 1538${}_{-51,+62}$
& 0.5
& 11.4\%
& 2040.22
& 1.22 \\

RealTabPFN-2.5 (D)
& 1510${}_{-44,+57}$
& 0.1
& 11.9\%
& 5.81
& 0.64 \\

RealMLP (T+E)
& 1486${}_{-44,+55}$
& 0.2
& 13.2\%
& 2950.72
& 11.97 \\

AutoGluon 1.4 (best, 4h)
& 1486${}_{-47,+54}$
& 0.0
& 13.1\%
& 1735.72
& 2.56 \\

TabDPT (T+E)
& 1441${}_{-48,+60}$
& 1.7
& 14.1\%
& 4910.38
& 286.54 \\

TabM (T+E)
& 1425${}_{-40,+49}$
& 0.2
& 14.5\%
& 3286.61
& 1.47 \\

LightGBM (T+E)
& 1411${}_{-31,+32}$
& 0.0
& 15.3\%
& 417.05
& 2.64 \\

RealMLP (T)
& 1410${}_{-46,+47}$
& 0.1
& 14.5\%
& 2950.72
& 0.66 \\

CatBoost (T+E)
& 1396${}_{-35,+40}$
& 0.1
& 14.9\%
& 1658.43
& 0.65 \\

TabDPT (T)
& 1388${}_{-54,+54}$
& 0.2
& 15.2\%
& 4910.38
& 39.96 \\

iLTM (T+E)
& 1388${}_{-43,+45}$
& 0.1
& 15.6\%
& 12685.08
& 464.37 \\

CatBoost (T)
& 1386${}_{-39,+39}$
& 0.4
& 15.1\%
& 1658.43
& 0.08 \\

TabM (T)
& 1373${}_{-41,+48}$
& 0.1
& 15.3\%
& 3286.61
& 0.17 \\

ModernNCA (T+E)
& 1370${}_{-53,+70}$
& 0.7
& 15.9\%
& 4621.67
& 8.14 \\

LightGBM (T)
& 1368${}_{-28,+29}$
& 0.0
& 15.9\%
& 417.05
& 0.33 \\

XGBoost (T+E)
& 1358${}_{-34,+32}$
& 0.0
& 16.0\%
& 693.49
& 1.69 \\

CatBoost (D)
& 1351${}_{-40,+36}$
& 0.0
& 15.8\%
& 6.83
& 0.08 \\

LimiX (D)
& 1347${}_{-60,+72}$
& 1.5
& 15.8\%
& 26.46
& 6.24 \\

ModernNCA (T)
& 1341${}_{-39,+39}$
& 0.3
& 16.3\%
& 4621.67
& 0.47 \\

XGBoost (T)
& 1336${}_{-33,+31}$
& 0.0
& 16.3\%
& 693.49
& 0.31 \\

xRFM (T+E)
& 1335${}_{-43,+46}$
& 0.0
& 16.6\%
& 846.89
& 2.55 \\

TabPFNv2 (T+E)
& 1328${}_{-65,+66}$
& 0.2
& 16.9\%
& 3031.50
& 21.44 \\

Mitra (D)
& 1316${}_{-65,+62}$
& 0.3
& 17.4\%
& 87.65
& 2.50 \\

TabDPT (D)
& 1314${}_{-54,+65}$
& 0.1
& 17.5\%
& 47.65
& 43.74 \\

TabICL (D)
& 1308${}_{-58,+50}$
& 0.0
& 17.3\%
& 6.63
& 1.48 \\

xRFM (T)
& 1291${}_{-39,+44}$
& 0.1
& 17.7\%
& 846.89
& 0.13 \\

TabM (D)
& 1286${}_{-42,+47}$
& 0.1
& 17.3\%
& 10.50
& 0.13 \\

iLTM (T)
& 1285${}_{-33,+38}$
& 0.1
& 17.4\%
& 12685.08
& 62.13 \\

TorchMLP (T+E)
& 1275${}_{-45,+49}$
& 0.0
& 17.4\%
& 2875.52
& 1.95 \\

SAP-RPT-OSS (D)
& 1272${}_{-54,+54}$
& 0.8
& 18.5\%
& 14.11
& 2.07 \\

TabPFNv2 (T)
& 1272${}_{-57,+58}$
& 0.2
& 18.4\%
& 3031.50
& 0.46 \\

BetaTabPFN (D)
& 1271${}_{-50,+55}$
& 0.1
& 18.8\%
& 205.88
& 1.34 \\

EBM (T+E)
& 1256${}_{-39,+38}$
& 0.0
& 18.9\%
& 2931.75
& 0.42 \\

TabPFNv2 (D)
& 1247${}_{-62,+64}$
& 0.3
& 19.1\%
& 3.36
& 0.31 \\

ModernNCA (D)
& 1240${}_{-38,+40}$
& 0.3
& 19.4\%
& 14.87
& 0.31 \\

EBM (T)
& 1222${}_{-42,+44}$
& 0.0
& 19.6\%
& 2931.75
& 0.05 \\

RealMLP (D)
& 1222${}_{-38,+35}$
& 0.1
& 18.8\%
& 10.06
& 1.69 \\

XGBoost (D)
& 1209${}_{-38,+36}$
& 0.0
& 19.0\%
& 1.94
& 0.12 \\

TorchMLP (T)
& 1204${}_{-44,+42}$
& 0.0
& 19.1\%
& 2875.52
& 0.13 \\

ExtraTrees (T+E)
& 1198${}_{-43,+42}$
& 0.1
& 20.2\%
& 183.02
& 0.76 \\

FastaiMLP (T+E)
& 1196${}_{-54,+49}$
& 0.2
& 19.9\%
& 593.24
& 4.47 \\

EBM (D)
& 1192${}_{-51,+46}$
& 0.1
& 20.5\%
& 7.33
& 0.05 \\

LightGBM (D)
& 1183${}_{-32,+31}$
& 0.0
& 19.6\%
& 1.96
& 0.14 \\

RandomForest (T+E)
& 1173${}_{-43,+51}$
& 0.1
& 21.1\%
& 373.24
& 0.77 \\

ExtraTrees (T)
& 1165${}_{-47,+42}$
& 0.1
& 21.1\%
& 183.02
& 0.09 \\

FastaiMLP (T)
& 1138${}_{-53,+50}$
& 0.1
& 21.3\%
& 593.24
& 0.31 \\

RandomForest (T)
& 1136${}_{-41,+50}$
& 0.2
& 21.8\%
& 373.24
& 0.09 \\

iLTM (D)
& 1089${}_{-52,+46}$
& 0.2
& 23.3\%
& 296.64
& 68.17 \\

TabSTAR (T)
& 1088${}_{-80,+74}$
& 0.9
& 25.8\%
& 28729.74
& 4.27 \\

TabSTAR (T+E)
& 1087${}_{-80,+78}$
& 1.1
& 25.8\%
& 28729.74
& 18.81 \\

OrionMSP (D)
& 1087${}_{-49,+52}$
& 0.0
& 23.9\%
& 13.12
& 2.52 \\

PerpetualBooster (T+E)
& 1087${}_{-46,+45}$
& 0.0
& 26.6\%
& 185.31
& 0.60 \\

TorchMLP (D)
& 1073${}_{-48,+35}$
& 0.1
& 23.0\%
& 9.99
& 0.13 \\

PerpetualBooster (T)
& 1050${}_{-44,+45}$
& 0.0
& 28.0\%
& 185.31
& 0.26 \\

xRFM (D)
& 1038${}_{-68,+58}$
& 0.0
& 26.7\%
& 3.23
& 0.92 \\

TabFlex (D)
& 1010${}_{-69,+60}$
& 0.1
& 27.8\%
& 0.79
& 0.12 \\

FastaiMLP (D)
& 1003${}_{-60,+55}$
& 0.1
& 25.9\%
& 2.86
& 0.37 \\

RandomForest (D)
& 1000${}_{-43,+41}$
& 0.0
& 26.5\%
& 0.43
& 0.05 \\

KNN (T+E)
& 990${}_{-83,+59}$
& 0.2
& 28.0\%
& 129.08
& 1.80 \\

TabSTAR (D)
& 988${}_{-96,+87}$
& 0.4
& 30.7\%
& 384.75
& 5.35 \\

ExtraTrees (D)
& 980${}_{-67,+55}$
& 0.0
& 27.8\%
& 0.25
& 0.05 \\

Linear (T+E)
& 956${}_{-101,+60}$
& 0.1
& 33.9\%
& 237.63
& 0.42 \\

Linear (T)
& 931${}_{-108,+67}$
& 0.0
& 34.4\%
& 237.63
& 0.08 \\

PerpetualBooster (D)
& 930${}_{-61,+42}$
& 0.1
& 31.9\%
& 27.32
& 0.03 \\

KNN (T)
& 885${}_{-95,+61}$
& 0.1
& 33.0\%
& 129.08
& 0.18 \\

Linear (D)
& 856${}_{-113,+68}$
& 0.1
& 37.0\%
& 1.19
& 0.12 \\

KNN (D)
& 644${}_{-90,+79}$
& 0.1
& 46.1\%
& 0.19
& 0.04 \\

\end{longtable}

\clearpage

\small
\setlength{\tabcolsep}{4.5pt}
\renewcommand{\arraystretch}{1.02}

\begin{longtable}{llccrr}

\caption{
Performance on the regression subset of TabArena, covering 13 datasets.
We compare models in terms of Elo, number of wins, improvability, and training and prediction time. \ours achieves the second-highest Elo while maintaining substantially lower computational cost than most tuned and ensembled baselines. Here \textbf{D} denotes the default (untuned) model, \textbf{T} the fine-tuned model, and \textbf{T+E} ensembling after fine-tuning.
}
\label{tab:elo_all} \\

\toprule
\textbf{Model}
& \textbf{Elo ($\uparrow$)}
& \textbf{\#wins ($\uparrow$)}
& \textbf{Improva-}
& \textbf{Train time}
& \textbf{Predict time} \\
&
&
&
\textbf{bility ($\downarrow$)}
& \textbf{per 1K [s]}
& \textbf{per 1K [s]} \\
\midrule
\endfirsthead

\multicolumn{6}{c}{
\textit{Table \thetable{} continued from previous page}
} \\
\toprule
\textbf{Model}
& \textbf{Elo ($\uparrow$)}
& \textbf{\#wins ($\uparrow$)}
& \textbf{Improva-}
& \textbf{Train time}
& \textbf{Predict time} \\
&
&
&
\textbf{bility ($\downarrow$)}
& \textbf{per 1K [s]}
& \textbf{per 1K [s]} \\
\midrule
\endhead

\midrule
\multicolumn{6}{r}{
\textit{Continued on next page}
} \\
\endfoot

\bottomrule
\endlastfoot

TABFM (D)
& \textcolor{gold}{\textbf{2019${}_{-127,+181}$}}
& \textcolor{gold}{\textbf{3.7}}
& \textcolor{gold}{\textbf{1.6\%}}
& 38.85
& 9.67 \\

\textbf{\ours (D)}
& \textcolor{silver}{\textbf{1900${}_{-148,+278}$}}
& \textcolor{silver}{\textbf{1.8}}
& \textcolor{silver}{\textbf{1.7\%}}
& 6.99
& 3.12 \\

EXAONE-Tabular (D)
& \textcolor{bronze}{\textbf{1885${}_{-110,+155}$}}
& \textcolor{bronze}{\textbf{1.5}}
& 3.0\%
& 13.71
& 2.58 \\

TabPFN-3 (D)
& 1800${}_{-131,+211}$
& 0.9
& \textcolor{bronze}{\textbf{2.4\%}}
& 3.87
& 0.42 \\

AutoGluon 1.5 (extreme, 4h)
& 1776${}_{-96,+137}$
& 1.3
& 3.9\%
& 335.03
& 4.33 \\

TabPFN-2.6 (D)
& 1741${}_{-56,+104}$
& 0.1
& 4.0\%
& 8.52
& 0.70 \\

RealTabPFN-2.5 (T+E)
& 1736${}_{-103,+153}$
& 0.2
& 3.4\%
& 1709.05
& 8.12 \\

TabDPT (T+E)
& 1722${}_{-90,+160}$
& \textcolor{bronze}{\textbf{1.5}}
& 4.3\%
& 4786.60
& 239.30 \\

TabICLv2 (D)
& 1679${}_{-142,+242}$
& 0.5
& 4.0\%
& 2.10
& 0.25 \\

TabDPT (T)
& 1670${}_{-73,+128}$
& 0.0
& 4.7\%
& 4786.60
& 38.50 \\

RealMLP (T+E)
& 1650${}_{-67,+111}$
& 0.1
& 5.1\%
& 3995.01
& 10.05 \\

RealTabPFN-2.5 (T)
& 1624${}_{-117,+154}$
& 0.2
& 4.1\%
& 1709.05
& 0.81 \\

AutoGluon 1.4 (best, 4h)
& 1592${}_{-90,+117}$
& 0.0
& 6.4\%
& 1866.35
& 6.07 \\

TabDPT (D)
& 1584${}_{-64,+137}$
& 0.0
& 5.6\%
& 46.62
& 39.21 \\

RealMLP (T)
& 1547${}_{-82,+105}$
& 0.0
& 6.0\%
& 3995.01
& 0.84 \\

RealTabPFN-2.5 (D)
& 1534${}_{-106,+141}$
& 0.0
& 5.6\%
& 7.04
& 0.51 \\

ModernNCA (T+E)
& 1531${}_{-118,+145}$
& 0.6
& 7.3\%
& 3779.70
& 7.69 \\

CatBoost (T+E)
& 1482${}_{-65,+103}$
& 0.0
& 8.0\%
& 3555.27
& 0.96 \\

LightGBM (T+E)
& 1474${}_{-83,+89}$
& 0.0
& 8.4\%
& 700.19
& 9.32 \\

xRFM (T+E)
& 1464${}_{-95,+112}$
& 0.0
& 7.5\%
& 714.50
& 1.38 \\

CatBoost (T)
& 1462${}_{-66,+102}$
& 0.0
& 8.1\%
& 3555.27
& 0.10 \\

TabM (T+E)
& 1442${}_{-85,+130}$
& 0.0
& 6.9\%
& 4160.58
& 1.41 \\

iLTM (T+E)
& 1438${}_{-46,+68}$
& 0.0
& 9.3\%
& 12685.08
& 321.74 \\

LightGBM (T)
& 1413${}_{-75,+91}$
& 0.0
& 9.0\%
& 700.19
& 0.97 \\

XGBoost (T+E)
& 1402${}_{-47,+64}$
& 0.0
& 9.0\%
& 834.93
& 2.61 \\

xRFM (T)
& 1384${}_{-81,+89}$
& 0.0
& 8.2\%
& 714.50
& 0.10 \\

XGBoost (T)
& 1382${}_{-53,+67}$
& 0.0
& 9.1\%
& 834.93
& 0.39 \\

CatBoost (D)
& 1369${}_{-89,+93}$
& 0.0
& 9.7\%
& 10.89
& 0.09 \\

ModernNCA (T)
& 1369${}_{-87,+105}$
& 0.0
& 9.3\%
& 3779.70
& 0.40 \\

TabM (T)
& 1366${}_{-97,+122}$
& 0.0
& 7.9\%
& 4160.58
& 0.17 \\

iLTM (T)
& 1333${}_{-57,+75}$
& 0.0
& 9.3\%
& 12685.08
& 59.10 \\

TabPFNv2 (T+E)
& 1318${}_{-116,+163}$
& 0.0
& 8.7\%
& 4223.87
& 27.54 \\

TabM (D)
& 1277${}_{-109,+113}$
& 0.0
& 9.5\%
& 13.32
& 0.13 \\

ModernNCA (D)
& 1274${}_{-68,+81}$
& 0.0
& 10.8\%
& 15.50
& 0.30 \\

Mitra (D)
& 1264${}_{-101,+126}$
& 0.1
& 10.3\%
& 71.06
& 1.85 \\

SAP-RPT-OSS (D)
& 1258${}_{-140,+144}$
& 0.1
& 10.8\%
& 20.24
& 6.62 \\

LimiX (D)
& 1246${}_{-154,+167}$
& 0.1
& 10.4\%
& 74.68
& 19.76 \\

TorchMLP (T+E)
& 1236${}_{-100,+110}$
& 0.0
& 11.0\%
& 4608.59
& 1.23 \\

TabPFNv2 (T)
& 1232${}_{-127,+153}$
& 0.0
& 9.7\%
& 4223.87
& 0.45 \\

RealMLP (D)
& 1226${}_{-91,+107}$
& 0.0
& 10.5\%
& 8.90
& 1.64 \\

ExtraTrees (T+E)
& 1210${}_{-94,+100}$
& 0.0
& 13.2\%
& 158.25
& 0.84 \\

LightGBM (D)
& 1206${}_{-41,+41}$
& 0.0
& 11.4\%
& 2.11
& 0.27 \\

XGBoost (D)
& 1189${}_{-77,+83}$
& 0.0
& 12.0\%
& 2.24
& 0.24 \\

TabPFNv2 (D)
& 1184${}_{-143,+133}$
& 0.0
& 11.1\%
& 2.80
& 0.31 \\

ExtraTrees (T)
& 1184${}_{-88,+92}$
& 0.0
& 13.4\%
& 158.25
& 0.15 \\

TorchMLP (T)
& 1180${}_{-96,+95}$
& 0.0
& 11.7\%
& 4608.59
& 0.10 \\

PerpetualBooster (T+E)
& 1180${}_{-85,+73}$
& 0.0
& 13.2\%
& 162.38
& 0.36 \\

RandomForest (T+E)
& 1162${}_{-61,+62}$
& 0.0
& 14.0\%
& 515.75
& 0.77 \\

xRFM (D)
& 1147${}_{-109,+110}$
& 0.0
& 13.8\%
& 2.45
& 0.74 \\

EBM (T+E)
& 1144${}_{-166,+124}$
& 0.0
& 14.5\%
& 2931.75
& 0.29 \\

PerpetualBooster (T)
& 1128${}_{-88,+57}$
& 0.0
& 14.3\%
& 162.38
& 0.17 \\

RandomForest (T)
& 1120${}_{-76,+63}$
& 0.0
& 14.5\%
& 515.75
& 0.12 \\

EBM (T)
& 1101${}_{-173,+126}$
& 0.0
& 15.0\%
& 2931.75
& 0.03 \\

ExtraTrees (D)
& 1069${}_{-107,+94}$
& 0.0
& 15.3\%
& 0.47
& 0.06 \\

EBM (D)
& 1042${}_{-173,+119}$
& 0.0
& 15.9\%
& 8.47
& 0.04 \\

TorchMLP (D)
& 1027${}_{-115,+81}$
& 0.0
& 15.1\%
& 20.49
& 0.08 \\

FastaiMLP (T+E)
& 1026${}_{-113,+103}$
& 0.0
& 15.3\%
& 540.14
& 2.67 \\



FastaiMLP (T)
& 979${}_{-111,+98}$
& 0.0
& 15.8\%
& 540.14
& 0.32 \\

TabSTAR (T+E)
& 951${}_{-284,+224}$
& 0.1
& 23.5\%
& 28729.74
& 19.90 \\

PerpetualBooster (D)
& 943${}_{-117,+79}$
& 0.0
& 17.7\%
& 27.32
& 0.02 \\

TabSTAR (T)
& 937${}_{-294,+225}$
& 0.1
& 23.7\%
& 28729.74
& 5.24 \\

KNN (T+E)
& 882${}_{-154,+140}$
& 0.0
& 21.0\%
& 92.55
& 0.90 \\

iLTM (D)
& 871${}_{-108,+67}$
& 0.0
& 19.0\%
& 357.02
& 81.61 \\

FastaiMLP (D)
& 864${}_{-160,+111}$
& 0.0
& 20.0\%
& 2.60
& 0.39 \\

TabSTAR (D)
& 832${}_{-345,+237}$
& 0.1
& 26.6\%
& 323.52
& 4.93 \\

KNN (T)
& 782${}_{-174,+145}$
& 0.0
& 23.3\%
& 92.55
& 0.05 \\

KNN (D)
& 680${}_{-246,+170}$
& 0.0
& 30.4\%
& 0.19
& 0.04 \\

Linear (T+E)
& 502${}_{-376,+127}$
& 0.0
& 37.4\%
& 193.98
& 0.17 \\

Linear (T)
& 469${}_{-447,+147}$
& 0.0
& 37.6\%
& 193.98
& 0.07 \\

Linear (D)
& 295${}_{-413,+146}$
& 0.0
& 40.0\%
& 0.95
& 0.10 \\

\end{longtable}

\clearpage

\section{Sensitivity Analysis}
\label{sec:noise-setup}

Although a number of mature benchmarks now enable systematic comparisons across different tabular foundation models, existing evaluation frameworks remain insufficient for fully characterizing how these models respond to external perturbations. To fill this gap, this section focuses on evaluating the robustness of tabular regression models to exogenous noise. In the overall factorial experimental design, we treat graph density and node-level signal-to-noise ratio (SNR) as two explicit and balanced experimental factors. For each matched base configuration, we fix the graph structure, structural functions, SNR vector, and feature permutation, and vary only the noise condition. Performance differences across noise conditions can therefore be attributed more directly to a model's sensitivity to exogenous noise, rather than to other uncontrolled variations in the data-generating process. The experimental results show that our proposed model maintains a substantial relative advantage across a wide range of noise distributions, SNR ranges, and graph density conditions, demonstrating strong robustness to exogenous perturbations.

\subsection{Experimental Design and Data Generation}

Each dataset is generated by a structural causal model with $32$ observed variables. Of these, $31$ variables serve as predictive features, and the last node in the sampled topological order serves as the regression target. No latent variables are introduced, so the evaluation is not additionally affected by unobserved variables or latent confounders. Although all variables are observable, the underlying graph structure, structural functions, and noise mechanisms remain unknown to the evaluated models.

Graphs are sampled using an ordered Erd\H{o}s--R\'enyi construction~\cite{Erdos}. Given a topological order, each candidate forward edge consistent with that order is added to the graph independently with probability $p$, so acyclicity is guaranteed by construction. Graph density is the first experimental factor and comprises two levels. In the sparse setting,
$p \sim \mathcal{U}[0.04,\,0.10]$, corresponding to an expected average in-degree of approximately $0.62$--$1.55$ and thus yielding relatively many root nodes. In the dense setting,
$p \sim \mathcal{U}[0.25,\,0.40]$, corresponding to an expected average in-degree of approximately $3.88$--$6.20$, with a substantially smaller number of root nodes. A clear gap is kept between the two intervals, so the sparse and dense settings form two clearly separated structural regimes.

We require the target variable to have at least one parent. If the initially sampled graph assigns no parent to the target variable, one to three nodes are randomly selected from those preceding the target in the topological order and connected to the target node. This correction rule guarantees that the target variable is structurally associated with at least one predictive feature.

For each base configuration, the structural function of every non-root node is sampled from a shared pool of random function generators. Function types and their parameters are sampled independently across nodes and across base configurations, but are held fixed across the fourteen experimental conditions associated with the same base configuration. Root nodes have no parents and are generated solely by exogenous random disturbances.

Let $\mathrm{pa}(j)$ denote the parent set of node $j$. All variables are generated sequentially in topological order. For root nodes, $x_j=\varepsilon_j$; for non-root nodes, generation follows
\begin{equation}
x_j
=
f_j\!\left(x_{\mathrm{pa}(j)}\right)
+
\sigma_j \varepsilon_j,
\qquad
\sigma_j = 10^{-s_j/20},
\label{eq:propagation}
\end{equation}
where $f_j$ denotes the sampled structural function, $\varepsilon_j$ denotes mutually independent exogenous noise terms, and $s_j$ denotes the SNR of node $j$ in decibels. For every non-root node, the signal term $f_j(x_{\mathrm{pa}(j)})$ and the raw noise term $\varepsilon_j$ are each standardized before being combined. Under this scheme, $\sigma_j=10^{-s_j/20}$ ensures that the SNR actually realized at the node equals the specified $s_j$ exactly. The resulting node value $x_j$ is then standardized again before being passed on to its descendants. Because child nodes receive the noisy $x_j$ rather than the denoised, purely structural signal, perturbations introduced at upstream nodes continue to propagate downward along the graph.

SNR is the second experimental factor and comprises four levels, denoted L1, L2, L3, and Lrand. Given an SNR level, each non-root node independently draws its own SNR value. For L1, L2, and L3, respectively,
\[
s_j \overset{\mathrm{i.i.d.}}{\sim} \mathcal{U}[-5,0), \qquad
s_j \overset{\mathrm{i.i.d.}}{\sim} \mathcal{U}[0,5), \qquad
s_j \overset{\mathrm{i.i.d.}}{\sim} \mathcal{U}[5,10].
\]
These three settings confine the node-level SNRs within a graph to narrow intervals of width $5$~dB. The three intervals are contiguous and of equal width, jointly forming an equal-width partition of the full range $[-5,10]$. For Lrand,
\[
s_j \overset{\mathrm{i.i.d.}}{\sim} \mathcal{U}[-5,10],
\]
so low-SNR and high-SNR nodes can coexist within the same graph. Lrand thereby treats within-graph SNR heterogeneity across nodes as an explicit experimental condition in its own right.

\subsection{Noise Diversity and Experimental Settings}

We evaluate fourteen experimental conditions in total (see Table~\ref{tab:noise}): twelve single-noise-distribution conditions, one mixed-noise condition, and one reference condition without additive noise. Under each single-distribution condition, all exogenous noise terms within a dataset are drawn from the same distribution family. Under the mixed condition, each node independently selects one of the twelve noise distribution families. Under the Clean reference condition, additive noise is removed from the structural equations of all non-root nodes by setting $\sigma_j=0$, while root nodes retain the exogenous random variation required to generate non-degenerate data.

All raw noise samples are standardized before entering the structural equations. Consequently, location and scale differences play no role when comparing the single-noise-distribution conditions; differences across conditions arise primarily from the post-standardization distributional shape and, where applicable, from randomly drawn shape parameters. The mixed condition further introduces heterogeneity in the noise distribution family across nodes.

The Exponential and Weibull conditions form a deliberately constructed, distributionally equivalent A/A comparison. A Weibull distribution with shape parameter $c=1$ is equivalent to an exponential distribution, and the rate parameter of the exponential distribution is eliminated by standardization. The two conditions therefore share the same standardized distribution at the population level, but are generated from mutually independent random samples. The difference between their results serves as an empirical yardstick for the performance fluctuation caused solely by finite-sample randomness, even when the noise laws are exactly equivalent.

\begin{table}[t]
\centering
\caption{The fourteen experimental conditions. Parameters specified by
$\mathcal{U}[\cdot]$ are independently resampled for each dataset.}
\label{tab:noise}
\small
\begin{tabular}{llll}
\toprule
Condition & Distribution & Parameters & Characteristic \\
\midrule
Normal      & $\mathcal{N}(0,1)$            & fixed                                    & symmetric, light-tailed \\
Laplace     & $\mathrm{Laplace}(0,1)$       & fixed                                    & symmetric, heavier-tailed \\
Uniform     & $\mathcal{U}[0,1]$            & fixed                                    & bounded support \\
Student-$t$ & $t_{\nu}$                     & $\nu \sim \mathcal{U}[2.1,\,5]$          & symmetric, heavy-tailed \\
Log-normal  & $\mathrm{LogNormal}(0,1)$     & fixed                                    & strongly right-skewed \\
Gumbel      & $\mathrm{Gumbel}(0,1)$        & fixed                                    & moderately right-skewed \\
Exponential & $\mathrm{Exp}(\lambda)$       & $\lambda \sim \mathcal{U}[0.5,\,2]$      & right-skewed \\
Chi-squared & $\chi^2_{k}$                  & $k \sim \mathcal{U}[2.1,\,5]$            & right-skewed \\
Beta        & $\mathrm{Beta}(\alpha,\beta)$ & $\alpha,\beta \sim \mathcal{U}[0.5,\,5]$ & bounded, flexible shape \\
Gamma       & $\mathrm{Gamma}(k,1)$         & $k \sim \mathcal{U}[0.5,\,5]$            & right-skewed \\
Half-normal & $|\mathcal{N}(0,1)|$          & fixed                                    & right-skewed \\
Weibull     & $\mathrm{Weibull}(c{=}1)$     & fixed                                    & A/A equivalent to exponential \\
Mixed       & per-node family choice        & one of the twelve families per node      & heterogeneous across nodes \\
Clean       & no non-root additive noise    & $\sigma_j=0$ for non-root $j$             & reference condition \\
\bottomrule
\end{tabular}
\end{table}

The experiment adopts a balanced full factorial design. Graph density has $2$ levels and SNR has $4$ levels, so crossing them yields $8$ experimental combinations. For each combination, we independently sample $20$ base configurations. Each base configuration comprises the sampled edge probability and the corresponding graph structure, the function type and function parameters of every non-root node, the node-level SNR vector, and a random permutation of the predictive feature columns.
Each base configuration is instantiated once under each of the fourteen experimental conditions, generating a total of $160 \times 14 = 2{,}240$
datasets.

\subsection{Experimental Results}
\label{sec:results-headtohead}

We compare \ours with Limix and TabICLv2. Across all $2{,}240$ datasets, \ours attains the highest $R^2$ $1{,}575$ times, corresponding to a first-place share of $70.3\%$; by comparison, the first-place shares of LimiX and TabICLv2 are only $16.0\%$ and $13.7\%$, respectively.

\ours's relative advantage remains stable across all thirteen non-Clean noise conditions. Its lowest first-place share is still $63.8\%$, a minimum attained under both the Laplace and Student-$t$ noise conditions, while its highest first-place share reaches $78.1\%$ under the Exponential noise condition. Even under its least favorable noise condition, \ours therefore still ranks first on nearly two-thirds of the datasets. In contrast, LimiX never exceeds a first-place share of roughly $28\%$ under any condition, with its maximum occurring under Student-$t$ noise, while TabICLv2's highest first-place share is roughly $25\%$, occurring under the Clean reference condition.

The distributionally equivalent exponential--Weibull pair serves as an internal A/A control. Their first-place shares differ by 5.0 percentage points despite having the same standardized population distribution, indicating that independent sample realizations alone can introduce noticeable variation. This pair therefore provides an empirical reference for the repeatability of the reported condition-wise results.

Under the Clean reference condition, \ours's first-place share is $65.6\%$, and under most noisy conditions its first-place share is no lower than this level. This indicates that \ours's relative competitive advantage does not depend on the absence of additive noise. By contrast, TabICLv2 attains its own highest first-place share under the Clean reference condition, and its relative competitiveness generally declines once additive noise is introduced. Overall, the experimental results show that \ours maintains a stable relative advantage across all tested noise distribution families and perturbation intensities.

\end{document}